\documentclass[wcp]{jmlr}

\usepackage{longtable}

\usepackage{booktabs}
\usepackage{hyperref}
\usepackage{url}
\usepackage{amsmath}
\usepackage{amssymb}
\usepackage{tikz}
\usetikzlibrary{positioning}
\usetikzlibrary{decorations.text}

\tikzset{anchor/.append code=\let\tikz@auto@anchor\relax}
\tikzset{my label/.style args={#1:#2}{
  append after command={
    (\tikzlastnode.center) node [#1] {#2}
    }
  }
}
\newcommand{\wog}{\color{blue} $w_{o_1}$}	
\newcommand{\wogg}{\color{orange} $w_{o_2}$}
\newcommand{\whg}{\color{red} $w_{h_1}$}
\newcommand{\whgg}{$w_{h_2}$}
\newcommand{\wng}{$w_{n_1}$}
\newcommand{\wngg}{$w_{n_2}$}
\newcommand{\ide}{$1$} 

\newcommand{\fname}{$f_1$} 
\newcommand{\fweight}{\color{cyan} $w_{f_1}$} 

\tikzstyle{edgenode}  =  [thin, draw=black, align=center,fill=white,font=\small]
\tikzstyle{edgeweight}  =  [near start, above, sloped,outer sep=3pt, inner sep=1pt ,fill=white,font=\large]
\tikzstyle{gredge}  =  [sloped,outer sep=2pt, inner sep=1pt ,fill=white]
\tikzstyle{gredge1}  =  [outer sep=2pt, inner sep=1pt ,fill=white]

\tikzstyle{kappa} = [
  draw, scale = 1.5,  minimum size=1cm, circle, rounded corners,shading=radial,inner color=white,
  minimum height=1cm,
  align=center
  ]
  
\tikzstyle{lambda} = [
  draw, scale = 1.5,  minimum size=1cm, circle, rounded corners,shading=radial,inner color=white,
  minimum height=1cm,
  align=center
  ]

\jmlrvolume{}
\jmlryear{}

\title[Lifted Relational Neural Networks]{Lifted Relational Neural Networks}

\author{\Name{Gustav \v{S}ourek} \Email{souregus@fel.cvut.cz}\\
\addr  Faculty of Electrical Engineering \\Czech Technical University in Prague \\Prague, Czech Republic
\AND
\Name{Vojt\v{e}ch Aschenbrenner} \Email{v@asch.cz}\\
\addr  Faculty of Mathematics and Physics \\Charles University \\Prague, Czech Republic
\AND
\Name{Filip \v{Z}elezn\'{y}} \Email{zelezny@fel.cvut.cz}\\
\addr  Faculty of Electrical Engineering \\Czech Technical University in Prague \\Prague, Czech Republic
\AND
\Name{Ond\v{r}ej Ku\v{z}elka}\thanks{Corresponding author.} \Email{KuzelkaO@cardiff.ac.uk}\\
\addr  School of CS \& Informatics \\Cardiff University \\Cardiff, United Kingdom}

\begin{document}

\maketitle

\begin{abstract}
We propose a method combining relational-logic representations with neural network learning. A general lifted architecture, possibly reflecting some background domain knowledge, is described through relational rules which may be handcrafted or learned. The relational rule-set serves as a template for unfolding possibly deep neural networks whose structures also reflect the structures of given training or testing relational examples. Different networks corresponding to different examples share their weights, which co-evolve during training by stochastic gradient descent algorithm. The framework allows for hierarchical relational modeling constructs and learning of latent relational concepts through shared hidden layers weights corresponding to the rules. Discovery of notable relational concepts and experiments on 78 relational learning benchmarks demonstrate favorable performance of the method.  
\end{abstract}

\section{Introduction}

{\em Lifted models} also known as \textit{templated} models have attracted significant attention recently \citep{LIFTED} in areas such as statistical relational learning. Lifted models define patterns from which specific (ground) models can be unfolded. For example, a lifted  Markov network model~\citep{MLN} may express that {\em friends of smokers tend to be smokers} and such a pattern then constrains the probabilistic relationships in all sets of vertices corresponding to particular friends-smokers in the derived ground Markov network. The lifted patterns are typically encoded in relational logic-based languages. 

Here we contribute a method for (deep) lifted feed-forward neural network learning, in which the ground network structure is unfolded from a set of weighted rules in relational logic. The relational rules are instantly interpretable and can be handcrafted by a domain expert or learned, e.g.\ through techniques of inductive logic programming \citep{Raedt}. Weights of the ground neural networks are determined by the weighted relational rules and can be learned by stochastic gradient descent algorithm. This means that weights between different ground neurons constructed from the same relational rule are tied in our framework, similarly to how weights are shared in lifted graphical models in statistical relational learning or how weights are tied together by application of filters in convolutional neural networks in deep learning.

A salient property of our approach distinguishing it from previous studies on adapting neural networks for relational learning is that the ground network structure depends not only on the relational rule set but also on a particular example, i.e., different networks are constructed for different examples to exploit their particular relational properties. However, the different networks share their weights as these are all bound to the relational rules, and so weight-updates performed for one training example are reflected in networks produced for other examples, which allows the model to learn directly from relational data.

The main advantage of the presented approach is that it can effectively learn weights of latent relational structures. This is a difficult task for existing lifted systems based on probabilistic inference because there one typically needs to run expensive expectation maximization algorithms in order to learn parameters when latent structures are present. On the other hand, deep neural networks, which we exploit in our work, have been shown to effectively learn latent structures, although obviously only in the ground non-relational settings. By combining relational logic with deep neural networks, we obtain a framework flexible enough to learn weights of latent relational structures, which we also verify experimentally. While there have been several works combining propositional or relational logic with neural networks \citep{KBANN,Botta,CILP}, none of the existing methods is able to learn weights of latent non-ground relational structures\footnote{What we mean by latent relational structures will be better explained in Section \ref{sec:concepts} where we present several types of latent structures which can be used in our framework.}.


The rest of the paper is organized as follows. The next section briefly summarizes the preliminaries regarding relational logic and the assumed neural network paradigm. Section \ref{sec:lrnns} explains the principles of the proposed Lifted Relational Neural Networks method. Section \ref{sec:concepts} describes useful modeling constructs. In Section \ref{sec:weightlearning}, we show how weight-learning is implemented in it. Section \ref{sec:related_work} places the presented methods in the context of existing works. In Section \ref{sec:experiments}, we subject the method to comparative experimental evaluation on relational learning benchmarks and then conclude the paper.

\section{Preliminaries}\label{sec:preliminaries}

A first-order logic theory is a set of formulas formed from constants, variables, functions, and predicates \citep{FOL}. Constant symbols represent objects in the domain of interest (e.g.\ $\textit{alice}$) and will be written in lower-case. Variables  (e.g.\ $\textit{Person}$) range over the objects in the domain and will be written with capitalized first letter. Function symbols will not be used in this paper. Predicate symbols represent relations among objects in the domain or their attributes. A {\em term} may be a constant or variable (or a function symbol applied to a tuple of terms). An {\em atom} is a predicate symbol applied to a tuple of terms (e.g.\ $\textit{friends}(X,\textit{bob})$). Formulas are constructed from atoms using logical connectives and quantifiers. A {\em ground term} is a term containing no variables. A {\em ground atom} is an atom having only ground terms as arguments (e.g.\ $\textit{friends}(\textit{alice},\textit{bob})$). A {\em literal} is an atom or a negation of an atom (which is also called a {\em negative literal}). A clause is a universally quantified disjunction of literals. When there is no risk of confusion, we will not write the universal quantifiers explicitly. A clause with exactly one positive literal is a {\em definite clause}. A definite clause with no negative literals (i.e. consisting of just one literal) is called a {\em fact}. A definite clause $h \vee \neg b_1 \vee \dots \vee \neg b_k$ can also be written as an implication $h \leftarrow b_1 \wedge \dots \wedge b_k$. The literal $h$ is then called {\em head} and the conjunction $b_1 \wedge \dots \wedge b_k$ is called {\em body}. We will sometimes call definite clauses, which are not facts, {\em rules}.

Given a first-order logic theory, the set of all ground atoms which can be constructed using the constants, function symbols and predicates present in the theory is its {\em Herbrand base}. 
A {\em Herbrand interpretation}, also called {\em possible world}, assigns a truth value to each possible ground atom from a given Herbrand base. A set of formulas is satisfiable if there exists at least one world in which all formulas from the set are true; such a world is its {\em Herbrand model}. A satisfiable set of definite clauses has a least Herbrand model and this model is unique. The least Herbrand model of a function-free set of definite clauses (i.e. a Datalog theory) can be constructed in finite number of steps using the {\em immediate-consequence operator}~\citep{ico}. Immediate consequence operator $T_p$ maps the space of Herbrand interpretations over some Herbrand base $\mathcal{B}$ back to itself as $ T_p : \mathcal{I}(\mathcal{B}) \mapsto \mathcal{I}(\mathcal{B})$. The~mapping of $T_p$ is directly prescribed by the theory $\mathcal{P}$ such that for $I \in \mathcal{I}(\mathcal{B})$ the $T_p(I) = \{h | (h \leftarrow b_1 \wedge \dots \wedge b_k) \in \mathcal{P}\}$ and $b_1 \wedge \dots \wedge b_k \subseteq I$. In other words the operator $T_p$ expands the current set of true atoms (interpretation $I$) with their immediate consequences as prescribed by the rules in $\mathcal{P}$.

An artificial neural network (NN) is a biologically inspired mathematical model, consisting of interconnected processing units called {\em neurons}, each of which is associated with an activation function $g_{i} \in \mathcal{G}$ from some predefined family of differentiable functions. Neural network then defines a mapping $f: \mathbb{R}^{m} \mapsto \mathbb{R}^{n}$ of input space to target space vectors, parameterized by a set of weights $w^l_j \in \mathbb{R}$. Following the pattern of neural interconnections, the mapping $f$ can be seen as a composition of activation functions $g_i \in G$. For feed forward neural networks it is typically a hierarchical compound of non-linear weighted sums $g_i(\sum_{j}w^l_j g_j(\sum_{k}w^{l+1}_k g_k(\ldots)))$, which can be conveniently depicted as a weighted directed acyclic graph of neurons (e.g.\ Fig~\ref{fig:ground}). By adapting the weights $w^i_j \in \mathcal{W}$ the model can be learned to approximate some target function $t : \mathbb{R}^{m} \mapsto \mathbb{R}^{n}$. This is typically performed by some sort of gradient descent minimization of a given cost function $cost : \{\mathcal{W},\mathcal{D}\} \mapsto \mathbb{R}$ capturing discrepancy between $f$ and $t$ upon some set of training samples $(x_d,t(x_d)) \in \mathcal{D}$. 

\section{Lifted Relational Neural Networks}\label{sec:lrnns}

A lifted relational neural network (LRNN) $\mathcal{N}$ is a set of weighted definite clauses, i.e. pairs $(R_i,w_i)$ where $R_i$ is a function-free definite clause and $w_i$ is a real number. When $\mathcal{N}$ is a set of weighted definite clauses, $\mathcal{N}^*$ will denote the corresponding set of the definite clauses without weights, i.e. $\mathcal{N}^* = \{ C : (C,w) \in \mathcal{N} \}$.  The set $\mathcal{N}$ must satisfy the following {\em non-recursiveness}\footnote{The reason why we do not allow recursion will be clearer when we explain weight learning in the next section. Here, we just note that whereas rule sets without recursion will lead to optimization problems solvable by an algorithm which is basically a modified back-propagation algorithm, rule sets with recursion  would lead to more complicated optimization problems which would not directly allow us to exploit existing results on training feedforward neural networks.} requirement: there must exist a strict ordering $\prec$ of predicates such that if there is a rule with a predicate $p_1$ in the head and a predicate $p_2$ in the body then $p_1 \prec p_2$. 

Given a LRNN $\mathcal{N}$, let $\mathcal{H}$ be the least Herbrand model of $\mathcal{N}^*$. We define {\em grounding of the LRNN $\mathcal{N}$}  as $\overline{\mathcal{N}} = \{ (h\theta \leftarrow b_1\theta \wedge \dots \wedge b_k\theta,w) : (h \leftarrow b_1 \wedge \dots \wedge b_k,w) \in \mathcal{N}\mbox{ and } \{ h\theta, b_1\theta, \dots, b_k\theta \} \subseteq \mathcal{H} \}$. That is, $\overline{\mathcal{N}}$ is defined as the set of ground definite clauses which can be obtained by grounding rules from the LRNN and which are active in the least Herbrand model of $\mathcal{N}^*$ (a rule is active in $\mathcal{H}$ if its body is true in $\mathcal{H}$). As already outlined in Introduction, LRNNs are templates for creating {\em ground} neural networks. The requirement that  ground rules should be active in $\mathcal{H}$ is beneficial for practice because it provides us with flexibility in controlling complexity of the constructed neural networks.

\begin{example}\label{example1}
Let
\begin{align*}
\mathcal{N} =& \{ (\textit{mother}(C,M) \leftarrow \textit{parent}(C,M) \wedge \textit{female}(M), 1), \\
& (\textit{father}(C,F) \leftarrow \textit{parent}(C,F) \wedge \textit{male}(F), 2),\\ & (\textit{female}(\textit{alice}),1), (\textit{parent}(\textit{bob},\textit{alice}), 1), (parent(\textit{eve},\textit{alice}),1) \}.
\end{align*}
Then for its grounding we have
\begin{align*}
\overline{\mathcal{N}} =& \{ (\textit{mother}(\textit{bob},\textit{alice}) \leftarrow \textit{parent}(\textit{bob},\textit{alice}) \wedge \textit{female}(\textit{alice}), 1), \\
&(\textit{mother}(\textit{eve},\textit{alice}) \leftarrow \textit{parent}(\textit{eve},\textit{alice}) \wedge \textit{female}(\textit{alice}), 1), \\
& (\textit{female}(\textit{alice}),1), (\textit{parent}(\textit{bob},\textit{alice}), 1), (\textit{parent}(\textit{eve},\textit{alice}),1) \}.
\end{align*}
Notice that $\overline{\mathcal{N}}$ does not contain the predicates $\textit{male}/1$ or $\textit{father}/2$ as there are no ground atoms based on them in the least Herbrand model of $\mathcal{N}$.
\end{example}

\begin{definition}
Let $\mathcal{N}$ be a LRNN, and let $\overline{\mathcal{N}}$ be its grounding. Let $g_{\vee}$, $g_{\wedge}$ and $g_{\wedge}^*$ be families of multivariate functions with exactly one function for each number of arguments. The {\em ground neural network} of $\mathcal{N}$ is a feedforward neural network constructed as follows.
\hfil
\begin{itemize}
\item For every ground atom $h$ occurring in $\overline{\mathcal{N}}$, there is a neuron $A_{h}$, called {\em atom neuron}. The activation functions of atom neurons are from the family $g_{\vee}$.

\item For every ground fact $(h,w) \in \overline{\mathcal{N}}$, there is a neuron $F_{(h,w)}$, called {\em fact neuron}, which has no input and always outputs a constant value.

\item For every ground rule $h\theta \leftarrow b_1\theta \wedge \dots \wedge b_k\theta \in \overline{\mathcal{N}}^*$, there is a neuron $R_{h\theta \leftarrow b_1\theta \wedge \dots \wedge b_k\theta}$, called {\em rule neuron}. It has the atom neurons $A_{b_1\theta}, \dots, A_{b_k\theta}$ as inputs, all with weight $1$. The activation functions of rule neurons are from the family $g_{\wedge}$.

\item For every rule $(h \leftarrow b_1 \wedge \dots \wedge b_k,w) \in \mathcal{N}$ and every $h\theta \in \mathcal{H}$, there is a neuron $\textit{Agg}_{(h \leftarrow b_1 \wedge \dots \wedge b_k,w)}^{h\theta}$, called {\em aggregation neuron}. Its inputs are all rule neurons $R_{h\theta' \leftarrow b_1\theta' \wedge \dots \wedge b_k\theta'}$ where $h\theta = h\theta'$ with all weights equal to $1$. The activation functions of the aggregation neurons are from the family $g_\wedge^*$.

\item Inputs of an atom neuron $A_{h\theta}$ are the aggregation neurons $\textit{Agg}_{(h \leftarrow b_1 \wedge \dots \wedge b_k,w)}^{h\theta}$ and fact neurons $F_{(h\theta,w)}$. The weights of the input neurons are the respective $w$'s.
\end{itemize}
\end{definition}

\begin{example}\label{ex:horses}
Let us consider the following LRNN
\begin{align*}
\mathcal{N} =& \{ (\textit{foal}(A) \leftarrow \textit{parent}(A,P) \wedge \textit{horse}(P), w_m), (\textit{foal}(A) \leftarrow \textit{sibling}(A,S) \wedge  \textit{horse}(S), w_n),\\
& (\textit{horse}(\textit{dakotta}), w_1), (\textit{horse}(\textit{cheyenne}), w_2), 
(\textit{horse}(\textit{aida}), w_3),\\
& (\textit{parent}(\textit{star},\textit{aida}), w_6),
(\textit{parent}(\textit{star},\textit{cheyenne}), w_5), (\textit{sibling}(\textit{star},\textit{dakotta}), w_4) \}.
\end{align*} 
The LRNN $\mathcal{N}$ and its ground neural network are shown in Fig. \ref{fig:family}.

\begin{figure}[h!]
\centering
\resizebox{1.0\textwidth}{!}{
	\begin{tikzpicture}
[node distance=2.5cm and 0.5cm,
mynode/.style={
  draw, scale = 1.7,  minimum size=1cm, circle, rounded corners,shading=radial,outer color=gray!30,inner color=white,
  minimum height=1cm,
  align=center
  },
myfact/.style={
  draw, scale = 1.7,  minimum size=1cm, rounded corners,left color=white,
  minimum height=1cm,
  align=center
  }
]

\begin{scope}[]

\begin{scope}[yshift=0.9cm, node distance=1.2cm and 0cm]
\node[myfact, label={[xshift=0.0cm, yshift=0.4cm]{\Large Facts}}, right color=black!20!white] (hd) {horse(dakota)};
\node[myfact, right color=black!20!white] (hc) [below of=hd] {horse(cheyenne)};
\node[myfact, right color=black!20!white] (ha) [below of=hc] {horse(aida)};

\node[myfact, right color=red!10!white] (msd) [below of=ha] {parent(star,aida)};
\node[myfact, right color=red!10!white] (fsc) [below of=msd] {parent(star,cheyenne)};
\node[myfact, right color=blue!10!white] (ssd) [below of=fsc] {sibling(star,dakotta)};

\end{scope}

\begin{scope}[xshift=6cm]
\node[mynode, label={[xshift=0cm, yshift=0.4cm]{\Large Rule-bodies}}, outer color=red!30!white] (mother) {parent(A,P)};
\node[mynode, outer color=black!30!white] (horse) [below of=mother]{horse(X)};
\node[mynode, outer color=blue!30!white] (father) [below of=horse] {sibling(B,S)}; 
\end{scope}

\begin{scope}[]
\node[mynode, label={[xshift=0cm, yshift=0.4cm]{\Large Rule-heads}}, outer color=violet!30!white] (foal) [right=2cm of horse] {foal(H)};
\end{scope}

\draw [thick,->] (mother) edge node[edgenode,sloped, near end] {$A=H$} node[gredge, near start, above,sloped] {1} (foal);
\draw [thick,->] (horse) edge node[edgenode,sloped, near end] {$ $} node[gredge, near start, above,sloped] {1} (foal);
\draw [thick,->] (father) edge node[edgenode,sloped, near end] {$B=H$} node[gredge, near start, above,sloped] {1} (foal);

\end{scope}

\begin{scope}[yshift=1.3cm, xshift=18cm]

\begin{scope}[xshift=-1cm,node distance=1.3cm and 0cm]
\node[myfact, label={[xshift=0.0cm, yshift=0.4cm]{\Large Fact neurons}}, right color=red!10!white] (fpsa) {parent(star,aida)};

\node[myfact, right color=black!20!white] (fha) [below of=fpsa] {horse(aida)};

\node[myfact, right color=red!10!white] (ffsc) [below of=fha] {parent(star,cheyenne)};
\node[myfact, right color=black!20!white] (fhc) [below of=ffsc] {horse(cheyenne)};

\node[myfact, right color=blue!10!white] (fssd) [below of=fhc] {sibling(star,dakotta)};
\node[myfact, right color=black!20!white] (fhd) [below of=fssd] {horse(dakotta)};

\end{scope}

\begin{scope}[xshift=6.5cm,node distance=1.3cm and 0cm]
\node[myfact, label={[xshift=0.0cm, yshift=0.4cm]{\Large Atoms neurons}}, my label={font = \Large, black,below=0.2cm:$\vee$}, right color=red!10!white] (apsa) {parent(star,aida)};

\node[myfact, my label={font = \Large, black,below=0.2cm:$\vee$}, right color=black!20!white] (aha) [below of=apsa] {horse(aida)};

\node[myfact, my label={font = \Large, black,below=0.2cm:$\vee$}, right color=red!10!white] (afsc) [below of=aha] {parent(star,cheyenne)};
\node[myfact, my label={font = \Large, black,below=0.2cm:$\vee$}, right color=black!20!white] (ahc) [below of=afsc] {horse(cheyenne)};

\node[myfact, my label={font = \Large, black,below=0.2cm:$\vee$}, right color=blue!10!white] (assd) [below of=ahc] {sibling(star,dakotta)};
\node[myfact, my label={font = \Large, black,below=0.2cm:$\vee$}, right color=black!20!white] (ahd) [below of=assd] {horse(dakotta)};

\end{scope}

\begin{scope}[yshift=-1.5cm, xshift=12.5cm, node distance=2.3cm and 0cm]
\node[myfact, label={[xshift=0.0cm, yshift=0.4cm]{\Large Rule neurons}}, right color=violet!30!white, my label={font = \Large, black,below=0.2cm:$\wedge$}] (mother) {foal(star)};
\node[myfact, right color=violet!30!white, my label={font = \Large, black,below=0.2cm:$\wedge$}] (father) [below of=mother] {foal(star)};
\node[myfact, right color=violet!30!white, my label={font = \Large, black,below=0.2cm:$\wedge$}] (sibling) [below of=father] {foal(star)};
\end{scope}

\begin{scope}[xshift=16.8cm, yshift=-2.8cm]
\node[myfact, label={[xshift=0.1cm, yshift=0.4cm]{\Large {Aggregation neurons}}},  right color=violet!30!white, my label={font = \Large, black,below=0.2cm:$\wedge*$}] (aggr1) {foal(star)};
\node[myfact, right color=violet!30!white, my label={font = \Large, black,below=0.2cm:$\wedge*$}] (aggr2) [below of=aggr1] {foal(star)}; {foal(star)};

\node[myfact, label={[xshift=0.0cm, yshift=0.4cm]{\Large Atom neuron}}, right color=violet!30!white, my label={font = \Large, black,below=0.2cm:$\vee$}] (final) [below right=0cm and 1.1cm of aggr1] {foal(star)};
\end{scope}

\draw [thick,->] (fhd) edge node[gredge, above,sloped] {$w_1$} (ahd);
\draw [thick,->] (fpsa) edge node[gredge, above,sloped] {$w_6$} (apsa);
\draw [thick,->] (fhc) edge node[gredge, above,sloped] {$w_2$} (ahc);
\draw [thick,->] (ffsc) edge node[gredge, above,sloped] {$w_5$} (afsc);
\draw [thick,->] (fssd) edge node[gredge, above,sloped] {$w_4$} (assd);
\draw [thick,->] (fha) edge node[gredge, above,sloped] {$w_3$} (aha);

\draw [thick,->] (apsa) edge node[gredge, above,sloped] {1} (mother);
\draw [thick,->] (aha) edge node[gredge, above,sloped] {1} (mother);

\draw [thick,->] (ahc) edge node[gredge, below, sloped] {1} (father);
\draw [thick,->] (afsc) edge node[gredge, above,sloped] {1} (father);

\draw [thick,->] (assd) edge node[gredge, below,sloped] {1} (sibling);
\draw [thick,->] (ahd) to[out=10,in=-150] node[outer sep=2pt, inner sep=1pt ,fill=white, below right] {1} (sibling);

\draw [thick,->] (mother) edge node[gredge, above,sloped] {1} (aggr1);
\draw [thick,->] (father) edge node[gredge, above,sloped] {1} (aggr1);

\draw [thick,->] (sibling) edge node[gredge, above,sloped] {1} (aggr2);

\draw [thick,->] (aggr1) edge node[gredge, above,sloped] {$w_m$} (final);
\draw [thick,->] (aggr2) edge node[gredge, above,sloped] {$w_n$} (final);

\end{scope}

\end{tikzpicture}
}
\caption{Depiction of the rule-based template (left) of LRNN $\mathcal{N}$ from Ex.~\ref{ex:horses}, and its corresponding ground neural network $\overline{\mathcal{N}}$ (right), with colors denoting the predicate signatures, rectangular nodes corresponding to ground and circular to lifted literals, respectively.}
\label{fig:family}
\end{figure}
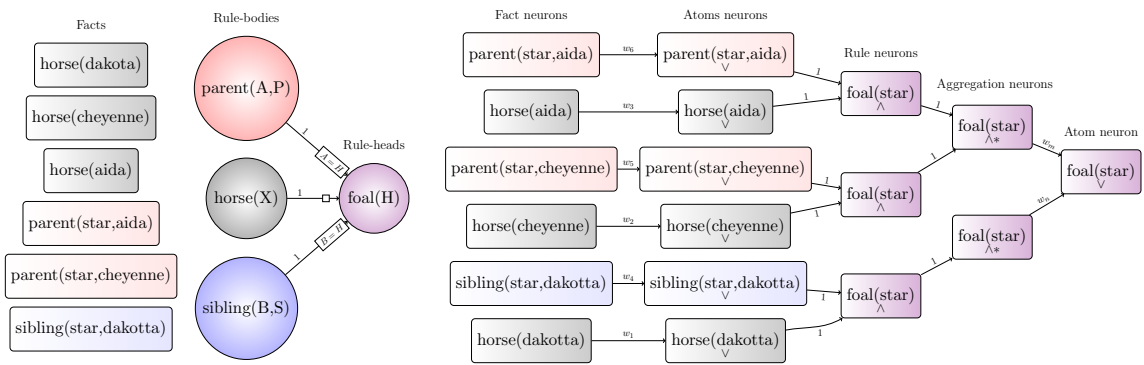

\end{example}

What distinguishes LRNNs from ordinary neural networks the most is the following property. Having a pre-trained LRNN $\mathcal{N}$ described by some general rules, we can extend it with description of a particular case to obtain a ground neural network and then use the latter for prediction. This is similar in spirit to lifted graphical models. 

\begin{example}

For instance, $\mathcal{N}$ may describe general rules for explosiveness of molecules (e.g.\ represented by a predicate $\textit{explosive}$) and $\mathcal{M}_1$ and $\mathcal{M}_2$ may be sets of (weighted) facts describing two particular molecules. Then to use the LRNN $\mathcal{N}$ for predicting whether $\mathcal{M}_1$ and $\mathcal{M}_2$ are explosive, we can simply construct ground NNs of $\overline{\mathcal{N}\cup \mathcal{M}_1}$ and $\overline{\mathcal{N}\cup \mathcal{M}_2}$, and compute the output of the respective atom neurons $\textit{explosive}^{1} \in \overline{\mathcal{N}\cup \mathcal{M}_1}$ and $\textit{explosive}^{2} \in \overline{\mathcal{N}\cup \mathcal{M}_2}$. 
As a distinctive feature of lifted models, the two ground LRNNs for the two example molecules may have very different size and structure because the least Herbrand models of $\mathcal{N}^* \cup \mathcal{M}_1^*$ and of $\mathcal{N}^* \cup \mathcal{M}_2^*$, which determine the structures of the ground LRNNs, may be very different (because the structure and the size of the molecules described by $\mathcal{M}_1$ and $\mathcal{M}_2$ are different). An illustration of this effect, for two example molecules and a template $\mathcal{N}$ from Fig.~\ref{fig:template}, is displayed in Fig.~\ref{fig:ground}.


\begin{figure}[t]
\centering
\resizebox{1.0\textwidth}{!}{
	\tikzstyle{atom}  =  [circle, scale=1.7, circle, shading=ball]
\newcommand\atoms{0}

\centering
\resizebox{1.0\textwidth}{!}{
\begin{tikzpicture}
\begin{scope}[yshift=1.5cm]
\node[atom, ball color=blue!50!white, label=left:$H(h^1)$, label=below left:{$bond(h^1,h^2)$}] at (0,6) (h3) {\textbf{H\textsuperscript{\tiny 1}}};
\node[atom, ball color=blue!50!white, label=right:$H(h^2)$, label=above right:{$bond(h^2,h^1)$}] at (1,5) (h4) {\textbf{H\textsuperscript{\tiny 2}}};

\node[atom, ball color=red!50!white, label=below left:{$bond(o^1,h^2)$}, label=above left:{$bond(o^1,h^1)$}, label=left:{$O(o^1)$}] at (0,1) (o1) {\huge \textbf{O\textsuperscript{{\tiny 1}}}};

\node[atom, circle, ball color=blue!50!white, label=right:{$H(h2)$}, label=below right:{$bond(h^2,o^1)$}] at (1,0) (h1) {\textbf{H\textsuperscript{\tiny 1}}};
\node[atom, circle, ball color=blue!50!white, label=right:{$H(h^1)$}, label=above right:{$bond(h^1,o^1)$}] at (1,2) (h2) {\textbf{H\textsuperscript{\tiny 2}}};

\end{scope}

\begin{scope}[xshift=7cm]
[node distance=2.5cm and 1cm]

\tikzstyle{neuroatom}  =  [circle, scale=1.6, circle, shading=radial, inner color = white, inner sep=0pt]

\newcommand\net{0}

\path [postaction={decorate,decoration={raise=-2pt,text along path,reverse path=true,text align/left indent={5.0cm},text=Atom-types}}] (0,7.0) circle (1.4cm);

\path [postaction={decorate,decoration={raise=-2pt,text along path,reverse path=true,text align/left indent={5.8cm},text=group-types}}] (6.1-.1,7.0) circle (1.4cm);
\path [postaction={decorate,decoration={raise=-2pt,text along path,reverse path=true,text align/left indent={5cm},text=atom-atom-bond}}] (6,4.0) circle (1.4cm);

\path [postaction={decorate,decoration={raise=-2pt,text along path,reverse path=true,text align/left indent={5.7cm},text=graphlet-features}}] (11.1-.1,5.65) circle (1.45cm);

\path [postaction={decorate,decoration={raise=-2pt,text along path,reverse path=true,text align/left indent={6cm},text=output-value}}] (15.15-.1,4.2) circle (1.4cm);

\node[lambda, outer color=red!30!white] at (0,\net + 7) (oo1) {O$(X_1)$};
\node[lambda, outer color=blue!30!white] at (0,\net + 1) (hh1) {H$(X_n)$};
\node[scale=2] at (0,\net + 3) (nic) {\huge \vdots};
\node[lambda, outer color=green!30!white] at (0,\net + 4.7) (nn1) {N$(X_2)$};

\node[kappa, outer color=brown!30!white] at (6,\net + 7) (gr1) {gr$_1(A)$};
\node[lambda, outer color=black!30!white] at (6,\net + 4) (bond) {b$(A,B)$};
\node[kappa, outer color=brown!30!white] at (6,\net + 1) (gr2) {gr$_2(B)$};

\node[lambda, outer color=violet!30!white] at (11,\net + 5.5) (grap) {\fname$(A,B)$};
\node[scale=1.8] at (11,\net + 3.85) (nic) {\huge \vdots};
\node[scale=.9, lambda, dashed, outer color=white!30!white] at (11,\net + 2) (empty) {$f_m(B,A)$};

\node[kappa, outer color=orange!30!white] at (15,\net + 4) (expl) {explosive};

\draw [thick,->] (oo1) edge node[edgenode,sloped, near end] {$X_1=A$} node[edgeweight] {\wog} (gr1);
\draw [thick,->] (hh1) edge node[edgenode,sloped, near end] {$X_n=A$} node[edgeweight] {\whg} (gr1);
\draw [thick,->] (nn1) edge node[edgenode,sloped, near end] {$X_2=A$} node[pos=0.32, below, sloped] {\wng} (gr1);

\draw [thick,->] (oo1) edge node[pos=0.15, above, sloped] {\wogg} node[edgenode,sloped, near end] {$X_1=B$} (gr2);
\draw [thick,->] (hh1) edge node[edgeweight] {\whgg} node[edgenode,sloped, near end] {$X_n=B$} (gr2);
\draw [thick,->] (nn1) edge node[edgeweight] {\wngg} node[edgenode,sloped, near end] {$X_2=B$} (gr2);

\draw [dashed,->] (gr1) edge node[near start, sloped, above] {1} (empty);
\draw [dashed,->] (bond) edge node[very near start, sloped, above] {1} (empty);
\draw [dashed,->] (gr2) edge node[near start, sloped, above] {1} (empty);

\draw [thick,->] (gr1) edge node[edgenode,sloped] {$A=A$} node[very near start, sloped, above] {1} (grap);
\draw [thick,->] (bond) edge node[edgenode,sloped] {$A=A$\\$B=B$} node[very near start, sloped, above] {1} (grap);
\draw [thick,->] (gr2) edge node[edgenode,sloped] {$B=B$} node[near start, sloped, above] {1} (grap);

\draw [thick,->] (grap) edge node[near start, sloped, above] {\fweight} (expl);
\draw [dashed,->] (empty) edge node[near start, sloped, above] {$w_{f_m}$} (expl);

\end{scope}
\end{tikzpicture}}
}

\caption{Two example molecules (left), described by surrounding sets of ground facts $\mathcal{M}_1$ and $\mathcal{M}_2$, are being merged with the lifted LRNN $\mathcal{N}$, composed of general weighted rules loosely pointing to explosiveness of molecules (right), to form two ground networks displayed in Fig.~\ref{fig:ground}. The rules in $\mathcal{N}$ provide adaptive means to create latent groups ($gr_i$) of atom types ($O\ldots H$) that, through a bond predicate ($b(A,B)$) connecting couples of atoms, form relational features (e.g.\ $f_1(A,B) \leftarrow gr_1(A) \wedge bond(A,B) \wedge gr_2(B)$), which set the basis for the final explosiveness output. For the sake of space we assume a single relational (graphlet) feature $f_1$ only.}
\label{fig:template}
\end{figure}
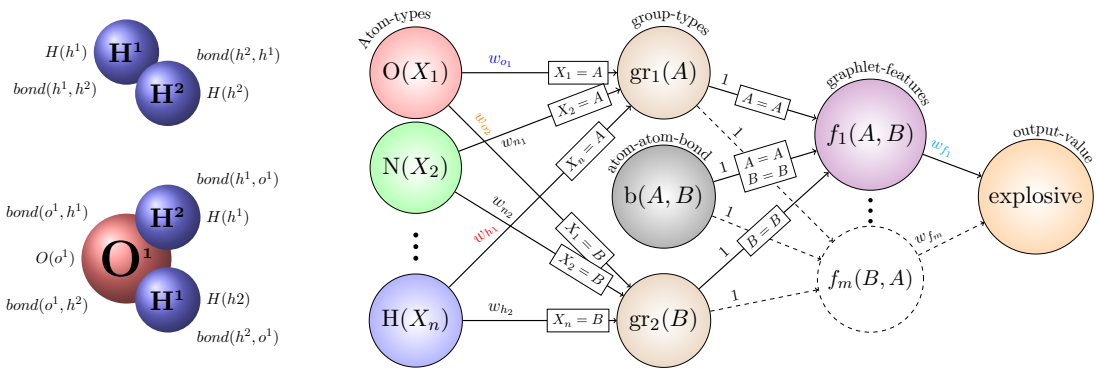

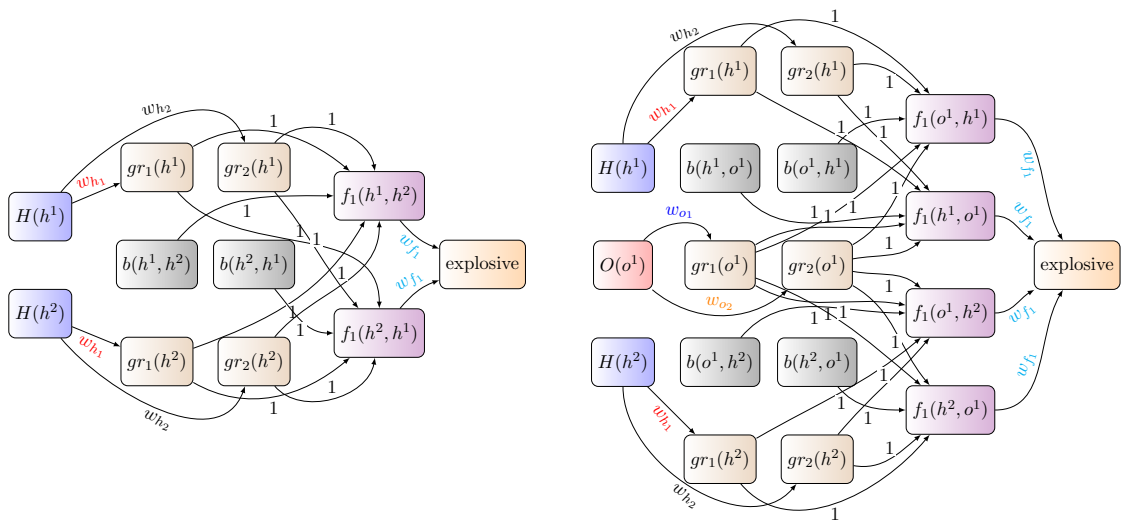
\begin{figure}[t]
\centering
\resizebox{1.0\textwidth}{!}{
	\centering
\resizebox{1.0\textwidth}{!}{
\begin{tikzpicture}
[node distance=2.0cm and 0.0cm,
ar/.style={->,>=latex},
mynode/.style={
  draw, scale = 1.0,  minimum size=1cm, rounded corners,left color=white,
  minimum height=1cm,
  align=center
  }
]

\tikzstyle{grbond}  =  [mynode, right color=black!30!white]
\tikzstyle{gratom}  =  [mynode]
\tikzstyle{grgroup} =  [mynode, right color=brown!30!white]
\tikzstyle{grexpl}  =  [mynode, right color=violet!30!white]
\tikzstyle{edgenode}  =  [thin, draw=black, align=center,fill=white,font=\small]

\newcommand\ground{15};
\coordinate (beg) at (5,5);

\begin{scope}[xshift=0cm,yshift=-1.0cm]
\begin{scope}[xshift=0cm]

\node[gratom, right color=blue!30!white] (gh1) {$H(h^1)$};
\node[gratom, right color=blue!30!white] (gh2) [below of=gh1] {$H(h^2)$};

\end{scope}

\begin{scope}[xshift=0cm]

\node[grgroup] (gr1h1) [above right=0.0cm and 1cm of gh1] {$gr_1(h^1)$};
\node[grgroup] (gr2h1) [right of=gr1h1] {$gr_2(h^1)$};

\node[grbond] (bondh1h2) [below of=gr1h1] {$b(h^1,h^2)$};
\node[grbond] (bondh2h1) [right of=bondh1h2] {$b(h^2,h^1)$};

\node[grgroup] (gr1h2) [below of=bondh1h2] {$gr_1(h^2)$};
\node[grgroup] (gr2h2) [right of=gr1h2] {$gr_2(h^2)$};
\end{scope}

\begin{scope}[xshift = 0cm]
\node[grexpl] (explosive1)  [above right=0.4 and 0.8cm of bondh2h1] {\fname($h^1,h^2$)};
\node[grexpl] (explosive2)  [below right=0.4 and 0.8cm of bondh2h1] {\fname($h^2,h^1$)};
\end{scope}
\node[grexpl, right color=orange!30!white] (expl) [below right=0.4 and 0.3cm of explosive1] {explosive};

\draw[ar] (gh1) -- node[gredge, above] {\whg} (gr1h1);
\draw[ar] (gh1) to[out=50,in=110] node[gredge, above] {\whgg} (gr2h1);

\draw[ar] (gh2) -- node[gredge, below] {\whg} (gr1h2);
\draw[ar] (gh2) to[out=-50,in=-110] node[gredge, below] {\whgg} (gr2h2);

\draw[ar] (gr1h1) to[out=30,in=140] node[gredge1, above] {\ide} (explosive1);
\draw[ar] (gr2h1) to[out=50,in=100] node[gredge1, above] {\ide} (explosive1);
\draw[ar] (gr1h2) to[out=20,in=-120] node[gredge1, below] {\ide} (explosive1);
\draw[ar] (gr2h2) to[out=50,in=-90] node[gredge1, above] {\ide} (explosive1);
\draw[ar] (bondh1h2) to[out=50,in=180] node[gredge1, below] {\ide} (explosive1);
\draw[ar] (gr1h1) to[out=-50,in=90] node[gredge1, above] {\ide} (explosive2);
\draw[ar] (gr2h1) to[out=-50,in=130] node[gredge1, above] {\ide} (explosive2);
\draw[ar] (gr1h2) to[out=-30,in=-140] node[gredge1, below] {\ide} (explosive2);
\draw[ar] (gr2h2) to[out=-50,in=-100] node[gredge1, above] {\ide} (explosive2);
\draw[ar] (bondh2h1) to[out=-50,in=180] node[gredge1, below] {\ide} (explosive2);
\draw[ar] (explosive1) to[out=-50,in=160] node[gredge, below] {\fweight} (expl);
\draw[ar] (explosive2) to[out=50,in=-160] node[gredge, above] {\fweight} (expl);

\end{scope}


\begin{scope}[xshift=12cm]
\begin{scope}[xshift=0cm]

\node[gratom, right color=blue!30!white] (gh1) {$H(h^1)$};
\node[gratom, right color=red!30!white] (go1) [below of=gh1]{$O(o^1)$};
\node[gratom, right color=blue!30!white] (gh2) [below of=go1] {$H(h^2)$};

\end{scope}

\begin{scope}[xshift=0cm]
\node[grgroup] (gr1o1) [right of=go1] {$gr_1(o^1)$};
\node[grgroup] (gr2o1) [right of=gr1o1] {$gr_2(o^1)$};

\node[grbond] (bondh1o1) [above of=gr1o1] {$b(h^1,o^1)$};
\node[grbond] (bondo1h1) [right of=bondh1o1] {$b(o^1,h^1)$};

\node[grgroup] (gr1h1) [above of=bondh1o1] {$gr_1(h^1)$};
\node[grgroup] (gr2h1) [right of=gr1h1] {$gr_2(h^1)$};

\node[grbond] (bondo1h2) [below of=gr1o1] {$b(o^1,h^2)$};
\node[grbond] (bondh2o1) [right of=bondo1h2] {$b(h^2,o^1)$};

\node[grgroup] (gr1h2) [below of=bondo1h2] {$gr_1(h^2)$};
\node[grgroup] (gr2h2) [right of=gr1h2] {$gr_2(h^2)$};
\end{scope}

\begin{scope}[xshift = 0cm]
\node[grexpl] (explosive1)  [above right=0.0 and 1cm of bondo1h1] {\fname($o^1,h^1$)};
\node[grexpl] (explosive2)  [below of=explosive1] {\fname($h^1,o^1$)};
\node[grexpl] (explosive3)  [below of=explosive2] {\fname($o^1,h^2$)};
\node[grexpl] (explosive4)  [below of=explosive3] {\fname($h^2,o^1$)};
\end{scope}

\node[grexpl, right color=orange!30!white] (expl) [below right=0.0 and 0.8cm of explosive2] {explosive};

\draw[ar] (gh1) -- node[gredge,above] {\whg} (gr1h1);
\draw[ar] (gh1) to[out=90,in=130] node[gredge,above] {\whgg} (gr2h1);

\draw[ar] (go1) to[out=50,in=110] node[gredge,above] {\wog} (gr1o1);
\draw[ar] (go1) to[out=-40,in=-140] node[gredge,above] {\wogg} (gr2o1);

\draw[ar] (gh2) -- node[gredge,below] {\whg} (gr1h2);
\draw[ar] (gh2) to[out=-90,in=-130] node[gredge,below] {\whgg} (gr2h2);

\draw[ar] (gr1h1) to[out=50,in=130] node[gredge1,above] {\ide} (explosive1);
\draw[ar] (gr2h1) to[out=10,in=140] node[gredge1,below] {\ide} (explosive1);
\draw[ar] (gr1o1) to[out=20,in=-140] node[gredge1,right] {\ide} (explosive1);
\draw[ar] (gr2o1) to[out=30,in=-130] node[gredge1,above] {\ide} (explosive1);
\draw[ar] (bondo1h1) to[out=50,in=180] node[gredge1,above] {\ide} (explosive1);
\draw[ar] (gr1h1) to[out=-30,in=140] node[gredge1,above] {\ide} (explosive2);
\draw[ar] (gr2h1) to[out=-50,in=130] node[gredge1,above] {\ide} (explosive2);
\draw[ar] (gr1o1) to[out=30,in=-170] node[gredge1,above] {\ide} (explosive2);
\draw[ar] (gr2o1) to[out=10,in=-140] node[gredge1,above] {\ide} (explosive2);
\draw[ar] (bondh1o1) to[out=-50,in=180] node[gredge1,above] {\ide} (explosive2);
\draw[ar] (gr1h2) to[out=30,in=-140] node[gredge1,below] {\ide} (explosive3);
\draw[ar] (gr2h2) to[out=50,in=-130] node[gredge1,above] {\ide} (explosive3);
\draw[ar] (gr1o1) to[out=-30,in=170] node[gredge1,below] {\ide} (explosive3);
\draw[ar] (gr2o1) to[out=-10,in=140] node[gredge1,below] {\ide} (explosive3);
\draw[ar] (bondo1h2) to[out=50,in=180] node[gredge1,below] {\ide} (explosive3);
\draw[ar] (gr1h2) to[out=-50,in=-130] node[gredge1,below] {\ide} (explosive4);
\draw[ar] (gr2h2) to[out=-10,in=-140] node[gredge1,above] {\ide} (explosive4);
\draw[ar] (gr1o1) to[out=-20,in=140] node[gredge1,above] {\ide} (explosive4);
\draw[ar] (gr2o1) to[out=-30,in=130] node[gredge1,below] {\ide} (explosive4);
\draw[ar] (bondh2o1) to[out=-50,in=180] node[gredge1,below] {\ide} (explosive4);
\end{scope}

\begin{scope}[]
\draw[ar] (explosive1) to[out=0,in=120] node[gredge,below] {\fweight} (expl);
\draw[ar] (explosive2) to[out=0,in=150] node[gredge,above] {\fweight} (expl);
\draw[ar] (explosive3) to[out=0,in=-150] node[gredge,below] {\fweight} (expl);
\draw[ar] (explosive4) to[out=0,in=-120] node[gredge,above] {\fweight} (expl);
\end{scope}

\end{tikzpicture}}
}
\caption{Two groundings $\overline{\mathcal{N}\cup \mathcal{M}_1}$ and $\overline{\mathcal{N}\cup \mathcal{M}_2}$ formed by merging the two example molecules with the LRNN $\mathcal{N}$ from Fig.~\ref{fig:template}. The shared predicate signatures and weights tied by the template are denoted by colors. For the sake of space we display only ground rule sets instead of complete ground networks (i.e., fact and aggregation neurons are omitted), Fig.~\ref{fig:family} illustrates the (direct) correspondence of such a set to a full ground neural network.}
\label{fig:ground}
\end{figure}
\end{example}


Depending on the used families of activation functions $g_\vee$, $g_\wedge$ and $g_{\wedge}^*$, we can obtain neural networks with different behavior. For intuitiveness, in order for rules $(h \leftarrow b_1 \wedge \dots \wedge b_k,w)$ to behave similarly to ``if-then'' rules, we should prefer the outputs of rule neurons to be {\em high} (e.g.\ close to 1) if and only if all the inputs from the atom neurons corresponding to the literals from the body of the rule have high outputs. Similarly, we should prefer the output of the atom neurons, which should intuitively behave similarly to disjunction, to be high if and only if at least one of the rule neurons or fact neurons, which are inputs for the given atom neuron, has high output. Logical operators from various fuzzy logics \citep{fuzzyLogic} may serve as an inspiration for selecting suitable activation functions. 

\begin{example}
In Goedel fuzzy logic, conjunction $b_1 \wedge \dots \wedge b_k$, where $b_i$ are fuzzy logic literals, is given as $\min_{i} b_i$ and disjunction $b_1 \vee \dots \vee b_k$ is given as $\max_i b_i$. To emulate reasoning in Goedel logic, we could simply set
$
g_{\wedge}(b_1, \dots, b_k) = \min_i b_i,$ 
$g_{\wedge}^*(b_1, \dots, b_m) = \max_i b_i,$ and $
g_{\vee}(b_1, \dots, b_m) = \max_i b_i
$.
Here, the output of any rule neuron $R_{h \leftarrow b_1 \wedge \dots \wedge b_k}$ is the minimum value which makes the fuzzy truth value of the implication $h \leftarrow b_1 \wedge \dots \wedge b_k$ equal to $1$ in the Goedel fuzzy logic. Likewise, the output of any aggregation neuron is the minimum value which makes the fuzzy truth value of all the respective ground implications equal to $1$ simultaneously. This way, LRNNs can emulate fuzzy logic programming.
\end{example}



Next, we introduce two particular collections of activation functions inspired by fuzzy logic which will be used in the experiments (note that the activation functions shown in the above example would not be very suitable for gradient-based learning).

\begin{definition}[Max-Sigmoid Activation Functions]
The Max-Sigmoid (MS) collection of activation functions is composed of the following three families of functions:
$
g_{\wedge}(b_1, \dots, b_k) = \textit{sigm} \left( \sum_{i=1}^k b_i -k + b_0  \right),
g_{\wedge}^*(b_1, \dots, b_m) = \max_i b_i$, and 
$g_{\vee}(b_1, \dots, b_k) = \textit{sigm} \left( \sum_{i=1}^k b_i + b_0 \right)
$.
\end{definition}

\begin{figure}
\resizebox{\textwidth}{5.0cm}{
\begin{tikzpicture}
  \node (img1) {\includegraphics[width=\textwidth]{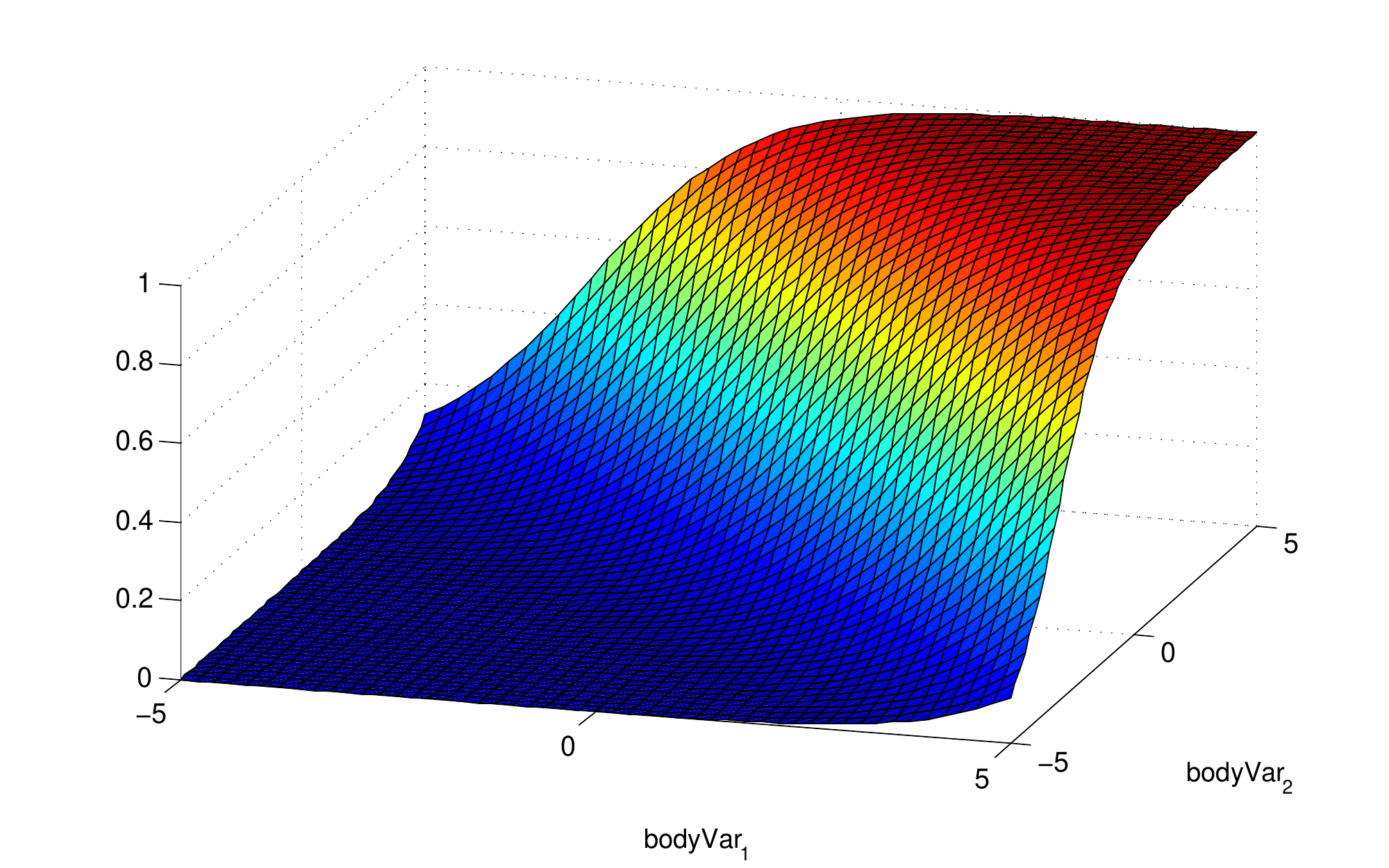}};
  \node (img3) at (img1.north west) [yshift=-1.5cm,xshift=4cm] {\includegraphics[width=4cm,height=8cm]{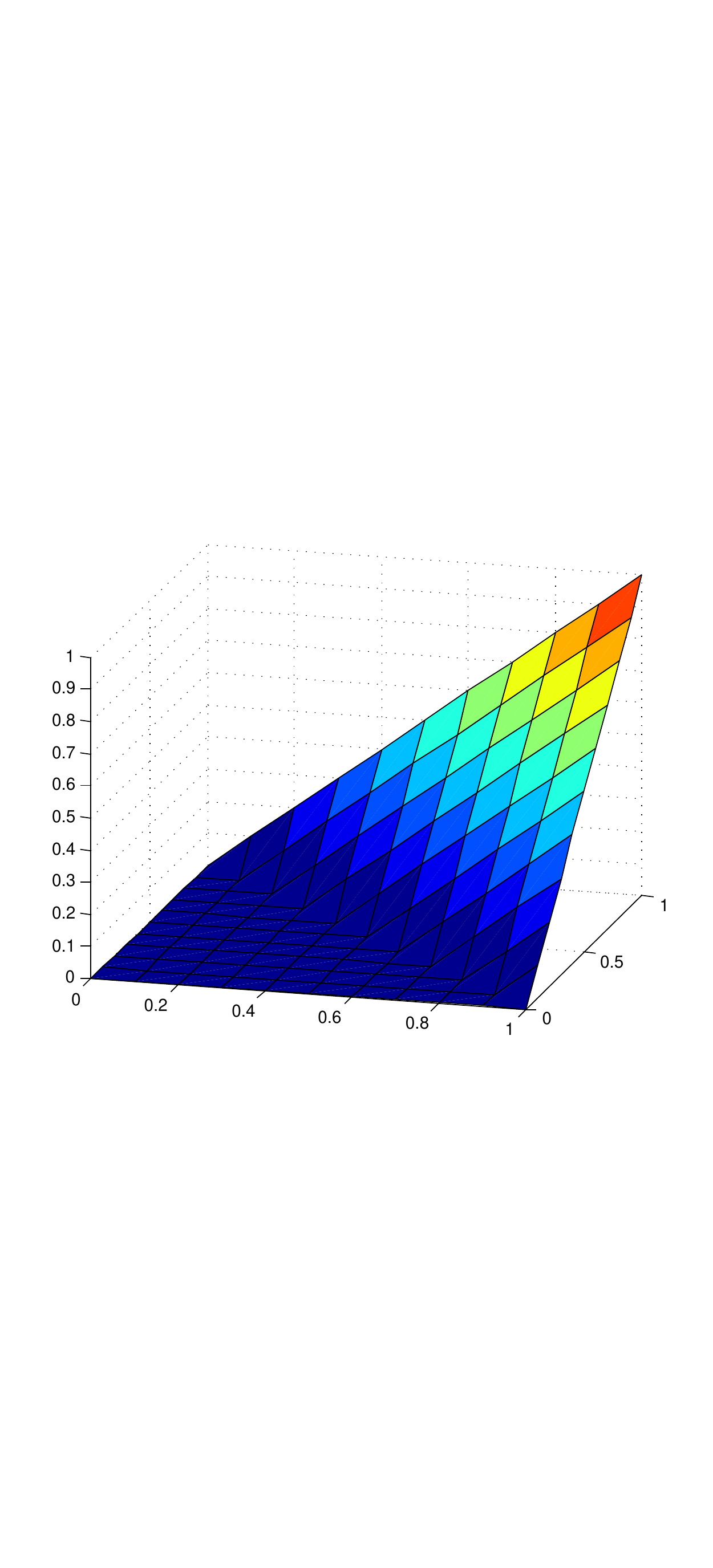}};
\end{tikzpicture}
\hspace{0.2cm}
\begin{tikzpicture}
  \node (img1) {\includegraphics[width=\textwidth]{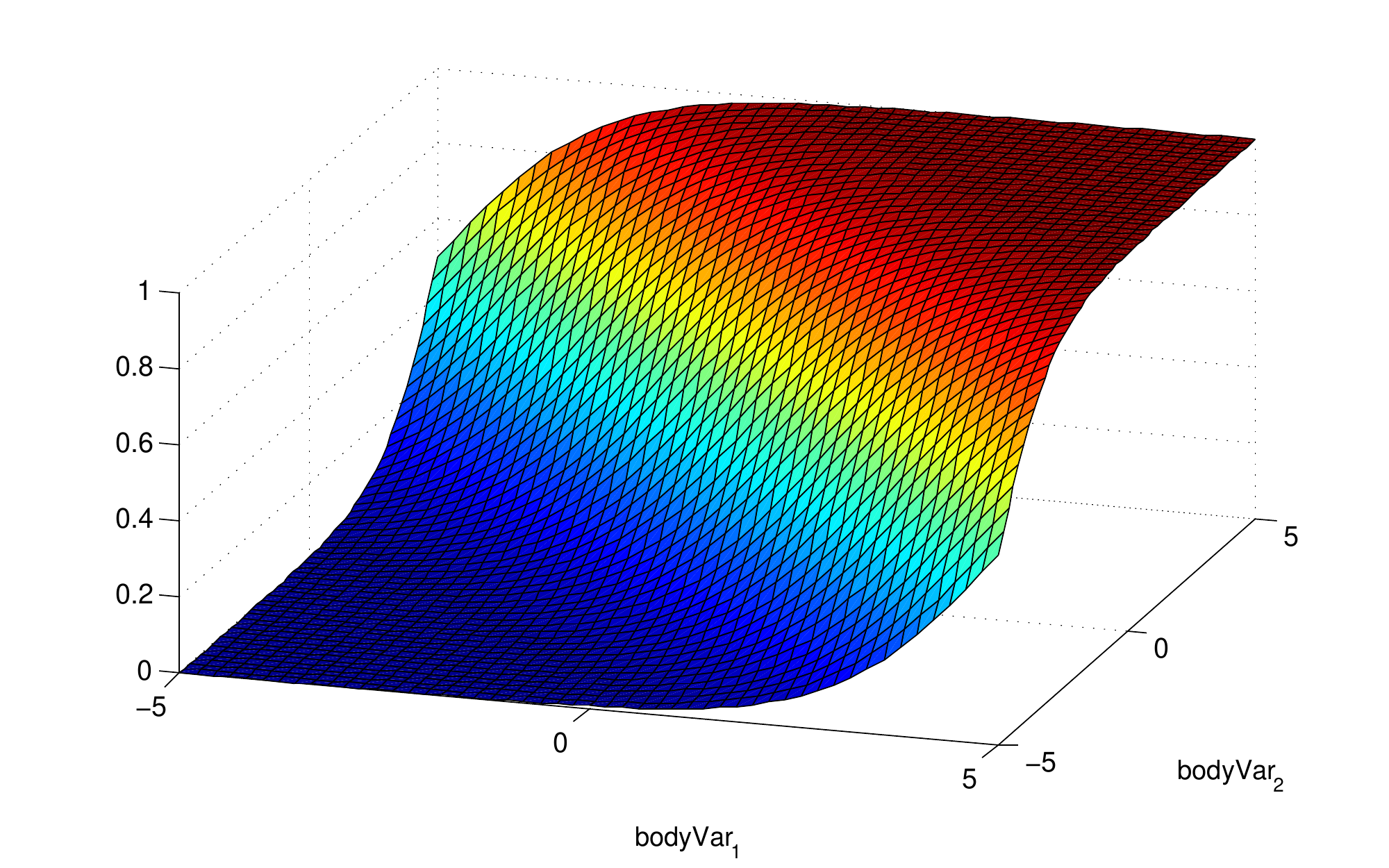}};
  \node (img3) at (img1.north west) [yshift=-1.5cm,xshift=3cm] {\includegraphics[width=4cm,height=8cm]{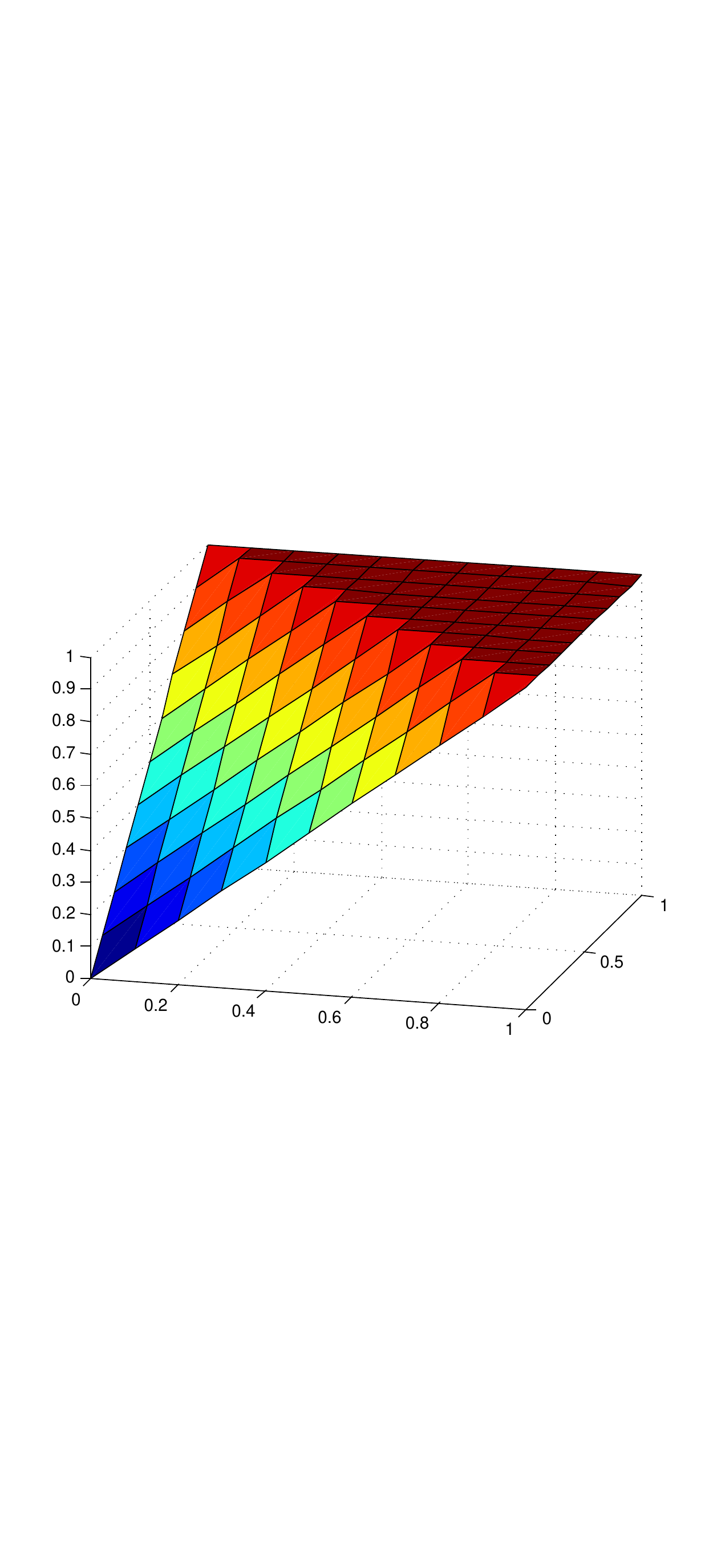}};
\end{tikzpicture}}
\caption{A crude approximation of Lukasiewicz conjunction (left) and disjunction (right) by respective sigmoidal activation functions for the use in LRNNs.}
\label{fig:sigmoid}
\end{figure}

The rationale for this family of activation functions is as follows. As already mentioned, the activation function $g_{\wedge}$ should have  high output if and only if all its inputs are high. To achieve this, we can crudely approximate Lukasiewicz fuzzy conjunction, which is given as $\max\{0, b_1 + \dots + b_k - k + 1 \}$, by the function $\textit{sigm} \left( b_1 + \dots + b_k -k + b_0 \right)$. A plot of the function $\textit{sigm} \left( b_1 + \dots + b_k -k + 1 \right)$ is shown, for $k = 2$, in the left panel of Fig.~\ref{fig:sigmoid}. The activation function $g_{\wedge}^*$ outputs the value equal to the highest of its inputs. Example~\ref{ex:max} illustrates that this can be seen as finding the best ``match'' of a pattern (rule). The activation function $g_\vee$ should have high output if at least one of the inputs is high or if all inputs are somewhat high. To satisfy this, we can crudely approximate Lukasiewicz fuzzy disjunction, which is given as $\min \{1, b_1 + \dots + b_k \}$ by the function $\textit{sigm} \left( b_1 + \dots + b_k + b_0 \right)$. A plot of the function  $\textit{sigm} \left( b_1 + \dots + b_k + 0\right)$ is shown in the right panel of Fig.~\ref{fig:sigmoid}. Example~\ref{ex:max2} illustrates the intuition for the activation function $g_\vee$.

\begin{example}\label{ex:max}
Let us consider the LRNN
\begin{align*}
\mathcal{N} =& \{ (\textit{hasBrightEdge} \leftarrow \textit{isBright}(E), 1),
(\textit{isBright}(E) \leftarrow \textit{edge}(E,U,V) \wedge \textit{bright}(U) \wedge \textit{bright}(V), \\  
& 1),  (\textit{bright}(U) \leftarrow \textit{yellow}(U), 2), (\textit{bright}(U) \leftarrow \textit{red}(U), 1), (\textit{bright}(U) \leftarrow \textit{blue}(U), 0.5) \}.
\end{align*}
Let us also have a set $\mathcal{G}$ describing a graph with colored vertices.
\begin{align*}
\mathcal{G} =& \{ (\textit{edge}(e_1,v_1,v_2), 1), (\textit{edge}(e_2,v_2,v_3), 1), (\textit{edge}(e_3,v_3,v_4), 1), (\textit{edge}(e_4,v_4,v_1), 1), \\
& (\textit{red}(v_1), 1), (\textit{blue}(v_2), 1), (\textit{yellow}(v_3), 1), (\textit{yellow}(v_4), 1)\}
\end{align*}
The output of the atom neuron $A_{\textit{hasBrightEdge}}$ will only depend on the ``brightest edge'', i.e. in this case on the edge $e_3$. The output would be the same for any other colored graph $\mathcal{G}'$, which would also contain an edge connecting two yellow vertices.
Thus, for instance, if we considered some physicochemical property of atoms (e.g.\ their partial charge) instead of brightness of colors, and molecules instead of colored graphs, the corresponding networks could detect presence of a molecular substructure similar to a prescribed pattern.
\end{example}

\begin{example}\label{ex:max2}
Let us have the LRNN
\begin{align*}
\mathcal{N} =& \{ (\textit{highPressure}(X) \leftarrow \textit{stressed}(X), 1), 
 (\textit{highPressure}(X) \leftarrow \textit{obese}(X), 1), \\
& (\textit{highPressure}(X) \leftarrow \textit{exercises}(X), -1) \}
\end{align*}
and the set of weighted facts $\mathcal{P} =$ $\{ (\textit{stressed}(\textit{alice}),1),$ $(\textit{obese}(\textit{alice}),1),$ $(\textit{stressed}(\textit{bob}),1),$ $(\textit{exercises}(\textit{bob}),1) \}.$
Outputs of aggregation neurons corresponding to rules from $\mathcal{N}$ with the same predicate in the head are combined using the activation functions $g_\vee$. Intuitively, rules and facts with the same predicate in the head can be seen as forming a logistic regression on the values given by the aggregation neurons from the lower layers. When the LRNN has just one layer, as in this example, one can achieve the same effect using techniques from {\em propositionalization}~\citep{propos} -- treating the bodies of the rules as features and feeding them as attributes to a logistic regression classifier. However, as soon as the LRNN has more layers, this effect cannot be emulated using propositionalization. In this particular example, if we construct the ground LRNN of $\overline{\mathcal{N}\cup \mathcal{P}}$ then the output of the atom neuron $A_{\textit{highPressure}(\textit{alice})}$ will be higher than the output of the atom neuron $A_{\textit{highPressure}(\textit{bob})}$ (because $\textit{alice}$ is stressed and obese whereas $\textit{bob}$ is just stressed and exercises).
\end{example}

The Max-Sigmoid activation function  is obviously not the only one possible. It is useful when we are interested in detecting one or more patterns (such as the existence of an edge as bright as possible in Example \ref{ex:max}) but less useful in situations similar to the one depicted in the next example.

\begin{example}
Let us consider the following simple LRNN for predicting individuals infected by flu
\begin{align*}
\mathcal{N} =&\{ (\textit{hasFlu}(A) \leftarrow \textit{friends}(A,B) \wedge \textit{hasFluDiagnosed}(B), 1) \}
\end{align*}
and a set of weighted ground facts $\mathcal{P}$ about a group of people and their friendships. If we constructed the ground neural networks of $\overline{\mathcal{N}\cup \mathcal{P}}$ using the activation functions from the Max-Sigmoid family then the prediction of whether an individual has flu would be entirely based on the existence of at least one person who already had flu diagnosed. It would be obviously more meaningful to base the predictions on the fraction of one's friends who had flu diagnosed.
\end{example}

A family of activation functions which are more appropriate in situations similar to to the one described in the above example is given by the next definition.

\begin{definition}[Avg-Sigmoid Activation Functions]
The Avg-Sigmoid (AS) collection of activation functions is composed of the following three families of functions:
$
g_{\wedge}(b_1, \dots, b_k) = \textit{sigm} \left( \sum_{i=1}^k b_i -k + b_0  \right), 
g_{\wedge}^*(b_1, \dots, b_m) = \frac{1}{m} \sum_{i=1}^m b_i$, and 
$g_{\vee}(b_1, \dots, b_k) = \sum_{i=1}^k b_i + b_0$.
\end{definition}

Another advantage of the Avg-Sigmoid family of activation functions over the Max-Sigmoid family is also that the functions from the Avg-Sigmoid family are everywhere differentiable (which simplifies learning). We note that other activation function families based on combinations of different aggregation functions might also be exploited for LRNN learning.

\section{Some LRNN Modeling Constructs}\label{sec:concepts}

In this section we describe several constructs which are easy in LRNNs but which would be difficult or impossible to implement in other existing frameworks combining logic and neural networks solely because, unlike LRNNs, the other frameworks do not allow simultaneous learning of target and auxiliary predicates. Moreover, while somewhat similar constructs could in principle be used in probabilistic logic programming systems such as Problog \citep{problog}, when learning, they would require running costly EM algorithms which repeatedly need to perform computationally expensive probabilistic inference.

\subsection{Implicit Soft Clustering}\label{sec:soft_clustering}

In many domains one needs to create clusters of certain objects in order to achieve good generalization. This is the case e.g.\ in prediction of adverse effects of drugs where significant improvements in predictive accuracy were gained by methods which were able to create auxiliary clusters of similar drugs \citep{jesse.adverseeffects}. However, the existing methods are still rather ad-hoc, relying on greedy discrete clustering. In LRNNs it is easy to define predicates representing these clusters, to train their weights automatically and use them for prediction of target predicates as illustrated by the following example.

\begin{example}
Let us suppose that, similarly to \citep{jesse.adverseeffects}, we have temporal data about patients, drugs which the patients took and time instants when changes in health occurred. Let us also assume that we have a set of general rules like:
$$
\begin{array}{lcl}
w_1^{(1)} : \textit{effect}(P,AE,T2) & \leftarrow & \textit{took}(P,D1,T1) \wedge \textit{period}(T1,T2,T) \wedge \textit{shortPeriod}(T) \wedge  \\
& & \wedge \textit{took}(P,D2,T2) \wedge \textit{drugGroup1}(D1) \wedge \textit{drugGroup2}(D2) \wedge \\
& & \wedge \textit{effectGroup1}(AE) \\
& & \dots \\
w_1^{(2)} : \textit{effectGroup1}(E) & \leftarrow & \textit{headache}(E) \\
w_2^{(2)} : \textit{effectGroup1}(E) & \leftarrow & \textit{sneezing}(E) \\
& \dots \\

\end{array}
$$
Using the Max-Sigmoid family of aggregation functions, weight learning in this LRNN can implicitly create clusters of drugs which interact adversely with other clusters of drugs and clusters of adverse effects corresponding to these combinations of drugs, as well as appropriate definition for the predicate $\textit{shortPeriod}$.
\end{example}
While we were not able to perform experiments in the domain described in the above example because the data are not available for privacy reasons, we perform a simpler set of experiments in organic chemistry domains where the implicitly created soft clusters correspond to groups of atom types and atomic bond types. We describe these experiments in detail in Section \ref{sec:experiments}. There we show that useful clusters are indeed created automatically by weight learning in LRNNs. One of the reasons for discussing the example about adverse effects of drugs here (in spite of the unavailability of the data) is to indicate that the machinery of LRNNs is very promising for existing problems for which only rather ad-hoc solutions exist currently.

\subsection{Soft Matching}\label{sec:soft_matching}

The next example explains the notion of a construct called {\em soft matching} and how it can be modeled in LRNNs.

\begin{example}
Let us again consider the example about predicting flu. Let us suppose that we have the reasonable rule that if $X$ is in a group of 4 people who are mutual friends and all of them have flu symptoms then $X$ has flu
$$\begin{array}{lcl}
w_1^{(1)} : \textit{hasFlu}(X) & \leftarrow & \textit{clique}(W,X,Y,Z) \wedge \textit{fluSymptoms}(W) \wedge \textit{fluSymptoms}(X) \wedge  \\
&  & \wedge \textit{fluSymptoms}(Y) \wedge \textit{fluSymptoms}(Z). \\
\end{array}$$
However, it is probably not necessary for $W$, $X$, $Y$ and $Z$ to be mutually friends in order for this rule to make sense. The rule is still valid, but maybe with lower certainty, if two of these four people are not actually friends, or maybe even if there are two such pairs or more. This is easily expressible in LRNNs by suitably defining the predicate $\textit{clique}$ and automatically learning the respective weights:
$$\begin{array}{lcl}
w_1^{(2)} : \textit{clique}(W,X,Y,Z) & \leftarrow & f(W,X) \wedge f(W,Y) \wedge f(W,Z) \wedge f(X,Y) \wedge f(X,Z) \wedge f(Y,Z) \\
w_1^{(3)} : \textit{f}(X,Y) & \leftarrow & \textit{friends}(X,Y) \wedge \textit{friends}(Y,X) \\
w_2^{(3)} : \textit{f}(X,Y) & \leftarrow & \textit{friends}(X,Y) \\
w_3^{(3)} : \textit{f}(X,Y). &  \\
\end{array}$$
Here, the predicate $\textit{friends}$ is assumed to be part of description of examples and soft matching of cliques is facilitated by the definition of the predicate $f$ based on it. Using the activation functions from the Max-Sigmoid family for the predicates $\textit{hasFlu}$ and $\textit{f}$, we can obtain the desired behavior with suitable weights.
\end{example}

\subsection{Other LRNN Concepts}\label{sec:other_concepts}

While soft clustering and soft matching are probably the modeling concepts which would be used most often in practice, there are other modeling concepts which are easily implementable with LRNNs. One such other concept is low dimensional approximation of sets of (hyper)graph patterns which share structure but not labels, as exemplified below.

\begin{example}\label{ex:other_concept}
Let us consider the problem of predicting a property, e.g.\ toxicity, of organic molecules which depends on presence of substructures from certain rather large set. If the patterns have the same structure, e.g.\ they are all aromatic six-rings with substitutions\footnote{The basic aromatic six-ring is the benzene ring which is a ring of six carbon atoms, each connected to a hydrogen atom, connected by aromatic bonds. If some of the carbon atoms is replaced by another atom, we speak of a {\em substitution}.} at some positions, one could in principle use probabilistic modeling to approximate this set by a probability distribution on the substitutions at different places so that the substitutions which are jointly occurring in the set of patterns would have high probability and the other substitutions small probability. While this probabilistic modeling approach is possible, it requires us to explicitly have the set of patterns. If the set of patterns should correspond to a latent concept, we would have to resort to EM. On the other hand, similar approximations to the latent set of patterns can be modeled in LRNNs quite easily. For instance, if we want to capture pair-wise dependencies of substitutions in neighboring atoms, we can first define auxiliary binary predicates
$$w_1^{(1)} : e_1(carbon, nitrogen), w_2^{(1)} : e_1(carbon, oxygen), \dots $$
Then, we can define a predicate
$$
w_1^{(2)} : sixRing(A,B,C,D,E,F) \leftarrow ring(A,B,C,D,E,F) \wedge e_1(A,B) \wedge e_2(B,C) \wedge \dots e_6(F,A)
$$
and similar predicates for five-rings and other structures, and then construct rules for prediction of the property of interest (toxicity in this case) as follows:
$$
\begin{array}{l}
w_1 : \textit{toxic}(M) \leftarrow \textit{atom}(M,A) \wedge \textit{atom}(M,B) \wedge \dots \wedge \textit{atom}(M,F) \wedge \textit{sixRing}(A,B,C,D,E,F) \\
w_2 : \textit{toxic}(M) \leftarrow \textit{atom}(M,A) \wedge \textit{atom}(M,B) \wedge \dots \wedge \textit{atom}(M,E) \wedge \textit{fiveRing}(A,B,C,D,E) \\
\dots
\end{array}
$$
Weight learning can then simultaneously adjust weights of the latent auxiliary predicates as well as the target predicates (we show this experimentally in Section \ref{sec:experiments}).
\end{example}

Exploiting the process of grounding of the lifted template, facilitating weight sharing in the ground networks, LRNNs can also emulate principal structures of convolutional neural networks \citep{CNNS} as the next example shows.

\begin{figure}[h!]
\centering
\resizebox{1.0\textwidth}{!}{
	\begin{tikzpicture}
[transform shape,rotate=0, node distance=2.0cm and 2.0cm,
ar/.style={->,>=latex},
mynode/.style={
  draw, scale = 1.0,  minimum size=1cm, rounded corners,left color=white,
  minimum height=1cm,
  align=center
  }
]

\tikzstyle{neuron}  =  [circle, draw, scale = 1.0,  minimum size=1cm]
\tikzstyle{grbond}  =  [mynode, right color=black!30!white]
\tikzstyle{gratom}  =  [mynode]
\tikzstyle{grgroup} =  [mynode, right color=brown!30!white]
\tikzstyle{grexpl}  =  [mynode, right color=violet!30!white]
\tikzstyle{edgenode}  =  [thin, draw=black, align=center,fill=white,font=\small]

\begin{scope}[xshift=0cm,yshift=0cm]

\begin{scope}[xshift=-1cm, yshift=-2cm]

\draw[step=2.0,black,thin] (0,0) grid (10,1.1);

\node at (1,0.5) {\Large $v_1$};
\node at (3,0.5) {\Large $v_2$};
\node at (5,0.5) {\Large $v_3$};
\node at (7,0.5) {\Large $v_4$};
\node at (9,0.5) {\Large $v_5$};

\node[] (gh2) [xshift=5cm, yshift=-0.7cm] {\LARGE image $I$ as a vector of pixel values};

\end{scope}

\begin{scope}[xshift=0cm]

\node[neuron,label={[xshift=-1.7cm, yshift=-0.5cm]{\Large features}},label={[xshift=-1.7cm, yshift=-1.2cm]{\Large (pixels)}}] (gh1) {$f_0^1$};
\node[neuron] (gh2) [right of=gh1] {$f_0^2$};
\node[neuron] (gh3) [right of=gh2] {$f_0^3$};
\node[neuron] (gh4) [right of=gh3] {$f_0^4$};
\node[neuron] (gh5) [right of=gh4] {$f_0^5$};

\end{scope}

\begin{scope}[xshift=1cm, yshift=2cm]

\node[neuron,label={[xshift=-2.2cm, yshift=-0.6cm]{\Large filter-map}},label={[xshift=-2.2cm, yshift=-1.2cm]{\Large (convolution)}}] (gr1h2) [above of=gh2] {$f_1$};
\node[neuron] (gr1h3) [right of=gr1h2] {$f_1$};
\node[neuron] (gr1h4) [right of=gr1h3] {$f_1$};

\end{scope}

\begin{scope}[xshift = 1cm, yshift=4cm]
\node[neuron, rectangle,label={[xshift=-2.0cm, yshift=-0.6cm]{\Large pooling}},label={[xshift=-2.0cm, yshift=-1.2cm]{\Large (max)}}] (explosive1)  [above of=gr1h3] {$pool_1$};
\end{scope}

\end{scope}


\draw[red,->] (gh1) -> node[red,near start,left] {$w_1$} (gr1h2);
\draw[red,->] (gh2) -> node[red,near start,left] {$w_1$} (gr1h3);
\draw[red,->] (gh3) -> node[red,near start,left] {$w_1$} (gr1h4);

\draw[blue,->] (gh2) -> node[blue] {$w_2$} (gr1h2);
\draw[blue,->] (gh3) -> node[blue] {$w_2$} (gr1h3);
\draw[blue,->] (gh4) -> node[blue] {$w_2$} (gr1h4);

\draw[green,->] (gh3) -> node[green,near start,right] {$w_3$} (gr1h2);
\draw[green,->] (gh4) -> node[green,near start,right] {$w_3$} (gr1h3);
\draw[green,->] (gh5) -> node[green,near start,right] {$w_3$} (gr1h4);

\draw[->] (gr1h2) -> node[left] {$1$} (explosive1);
\draw[->] (gr1h3) -> node[right] {$1$} (explosive1);
\draw[->] (gr1h4) -> node[right] {$1$} (explosive1);

\begin{scope}[xshift=11cm,yshift=0cm]

\begin{scope}[xshift=-1cm, yshift=-2cm]

\draw[step=2.0,black,thin,dashed] (0,0) grid (10,1.1);

\node at (1,0.8) {\large $(f_0(1),v_1)$};
\node at (3,0.8) {\large $(f_0(2),v_2)$};
\node at (5,0.8) {\large $(f_0(3),v_3)$};
\node at (7,0.8) {\large $(f_0(4),v_4)$};
\node at (9,0.8) {\large $(f_0(5),v_5)$};

\node[fill=white,inner sep=0pt,outer sep=0pt] at (1.3,0.25) {\large $(next(1,2),0)$ };
\node[fill=white,inner sep=0pt,outer sep=0pt] at (3.8,0.25) {\large $(next(2,3),0)$ };
\node[fill=white,inner sep=0pt,outer sep=0pt] at (6.25,0.25) {\large $(next(3,4),0)$ };
\node[fill=white,inner sep=0pt,outer sep=0pt] at (8.7,0.25) {\large $(next(4,5),0)$ };

\node[] (gh2) [xshift=5cm, yshift=-0.7cm] {\LARGE image $I$ as a set of weighted facts};

\end{scope}

\begin{scope}[xshift=0cm]

\node[gratom, right color=blue!30!white] (gh1) {$f_0(1)$};
\node[gratom, right color=blue!30!white] (gh2) [right of=gh1] {$f_0(2)$};
\node[gratom, right color=blue!30!white] (gh3) [right of=gh2] {$f_0(3)$};
\node[gratom, right color=blue!30!white] (gh4) [right of=gh3] {$f_0(4)$};
\node[gratom, right color=blue!30!white] (gh5) [right of=gh4] {$f_0(5)$};

\end{scope}

\begin{scope}[xshift=1cm, yshift=2cm]

\node[grbond] (gr1h2) [above of=gh2] {$f_1$};
\node[grbond] (gr1h3) [right of=gr1h2] {$f_1$};
\node[grbond] (gr1h4) [right of=gr1h3] {$f_1$};

\end{scope}

\begin{scope}[xshift = 1cm, yshift=4cm]
\node[grexpl] (explosive1)  [above of=gr1h3] {$Agg^{f_1}$};
\end{scope}


\draw[red,->] (gh1) -> node[red,near start,left] {$w_1$} (gr1h2);
\draw[red,->] (gh2) -> node[red,near start,left] {$w_1$} (gr1h3);
\draw[red,->] (gh3) -> node[red,near start,left] {$w_1$} (gr1h4);

\draw[blue,->] (gh2) -> node[blue] {$w_2$} (gr1h2);
\draw[blue,->] (gh3) -> node[blue] {$w_2$} (gr1h3);
\draw[blue,->] (gh4) -> node[blue] {$w_2$} (gr1h4);

\draw[green,->] (gh3) -> node[green,near start,right] {$w_3$} (gr1h2);
\draw[green,->] (gh4) -> node[green,near start,right] {$w_3$} (gr1h3);
\draw[green,->] (gh5) -> node[green,near start,right] {$w_3$} (gr1h4);

\draw[->] (gr1h2) -> node[left] {$1$} (explosive1);
\draw[->] (gr1h3) -> node[right] {$1$} (explosive1);
\draw[->] (gr1h4) -> node[right] {$1$} (explosive1);

\end{scope}

\end{tikzpicture}
}
\caption{A demonstration of a part of standard Convolutional Neural Network structure with sparse, convolutional layer composed of application of a filter $f_1$ creating a feature-map layer followed by max-pooling (left). The same calculation structure is presented by a ground LRNN (right) efficiently encoded with a template from Example~\ref{ex:CNN}, generalizing over feature-vectors (images) of unrestricted size.}
\label{fig:CNN}
\end{figure}
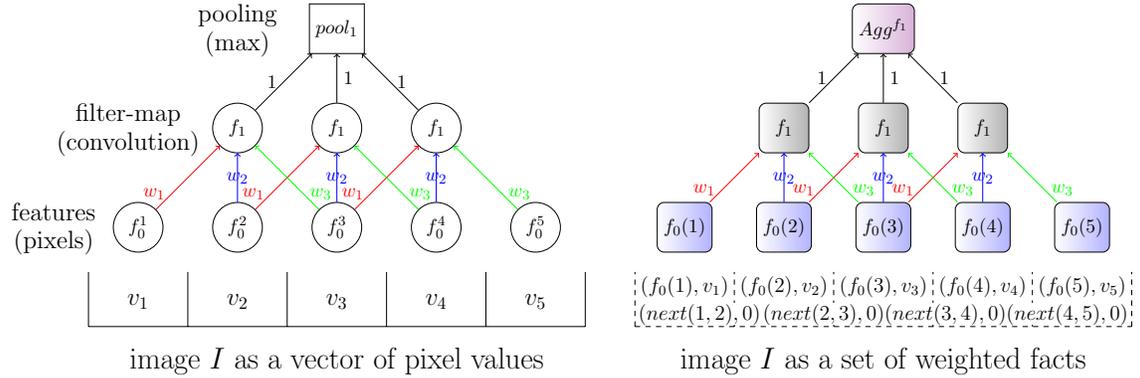

\begin{example}
\label{ex:CNN}

Let us consider a structure of the popular Convolutional Neural Network architecture composed of sparse convolutional layers alternated with max-pooling. Within the sparse layer, the weights corresponding to a single convolution filter are effectively bound to the same value while the filter is repeated across. Within selected subregions, the resulting feature-map values are then aggregated with application of max-pooling, i.e. only the maximal values from each feature-map region are propagated further. This structural idea can be efficiently encoded by LRNN and generalized for feature maps (images) of varying size with the choice of Max-Sigmoid function family and a simple lifted template defined as follows
$$
\begin{array}{llll}
w_0^{(1)} : f_1 & \leftarrow & left(A), mid(B), right(C),\; next(A,B), next(B,C) & \\
w_1^{(2)} : left(X) & \leftarrow & f_0(X) & \\ 
w_2^{(2)} : mid(X) & \leftarrow & f_0(X) & \\ 
w_3^{(2)} : right(X) & \leftarrow & f_0(X)
\end{array}
$$
which corresponds to a convolution filter $f_1$ that can be bound to an arbitrary number of relational patterns, in this case simple linear segments of three neighboring features ($A,B,C$), of the input feature-map defined as a linearly ordered set of weighted facts about feature $f_0$ values $(f_0(X),v_x)$ (i.e. values $v_x$ of pixels $X = \{1\ldots n\}$). The choice of Max-Sigmoid family then ensures max-aggregation to be applied on top of each such a convolutional layer. Visualization of a grounding of this template on a particular feature-map (image $I$) of five ($n=5$) consecutive values (pixels) is provided in Fig~\ref{fig:CNN}.

\end{example}

Other concepts which we do not describe in detail due to lack of space include e.g.\ relational auto-encoders.

\section{Weight Learning}\label{sec:weightlearning}

Let us have a LRNN $\mathcal{N}$ and a set of training examples $\mathcal{E} = \{\mathcal{E}^1, \dots, \mathcal{E}^m\}$ where each $\mathcal{E}^j$ is some structure represented by a set of weighted propositions (e.g.\ left part of Fig.~\ref{fig:template}), i.e. a LRNN containing only facts\footnote{The restriction of learning from facts only is actually not necessary but it will simplify this presentation.}. Let us also have a set $\mathcal{Q}$ $=$ $\{ \{ (q_1^1, t_1^1),$ $\dots,$ $(q_{k_1}^1,$ $t_{k_1}^1) \}$ $,\dots,$ $\{ (q_1^m, t_1^m),$ $\dots,$ $(q_{k_m}^m, t_{k_m}^m) \} \}$ where $q_i^j$ are ground atoms, which we call {\em training query atoms}, and $t_i^j$ are their {\em target values}. For any query atom $q_{i}^j$, let $y_i^j$ denote the output of the atom neuron $A_{q_i^j}$ in the ground neural network of $\overline{\mathcal{N}\cup \mathcal{E}^j}$. The goal of the learning process is to find weights $w_h$ of the rules (and possibly facts) in $\mathcal{N}$ minimizing cost $J$ on the training query atoms $J(\mathcal{Q}) = \sum_{j = 1}^m \sum_{i = 1}^{k_j} \textit{cost}(y_i^j,t_i^j)$ where $\textit{cost}$ is some predefined cost function which measures the discrepancy between the output of the atom neurons of the training query atoms and their desired target values.
Similarly to conventional NNs, weight adaptation is performed by gradient descent steps
$$ w_h \gets w_h - \gamma \frac{\partial J(\mathcal{Q})}{\partial w_h} $$
where $\gamma$ is some given learning rate.
The main difference is that in the case of LRNNs, the ground neural networks may be very different for different learning examples $\mathcal{E}^j$. However, this is not a fundamental problem because the weights for all the ground neural networks $\overline{\mathcal{N}\cup \mathcal{E}^j}$ are fully specified in the LRNN $\mathcal{N}$.

\begin{example}
Let us demonstrate for clarity a sample scenario with Avg-Sigmoid activation function family and a mean square error cost function, i.e.\ with each step we aim to decrease\footnote{In this example, we pass the output from the output atom neurons and the target values through a sigmoid. This is useful when learning with the Avg-Sigmoid activation function family. An alternative would be to use cross-entropy as error function.}
$$J(\mathcal{Q}) = \frac{1}{2} \sum_{j = 1}^m \sum_{i = 1}^{k_j} \left(\textit{sigm}(t_i^j) - \textit{sigm}(y_i^j)\right)^2$$
where the target values $t_i^j$ are given by $\mathcal{Q}$ and outputs $y_i^j$ of individual atom neurons $A_{q_i^j}$ are calculated as
$$A_{q_i^j} = \left( \sum_{k} w_k \: \textit{Agg}_{(h \leftarrow b_1 \wedge \dots \wedge b_{n_k},w_k)}^{q_i^j} \right) - w_{A_{q_i^j}} $$
where $\textit{Agg}_{(h \leftarrow b_1 \wedge \dots \wedge b_{n_k},w_k)}^{q_i^j}$ denotes the outputs of aggregation neurons forming the inputs of $A_{q_i^j}$ with respective rule weights $w_k$, and $w_{A_{q_i^j}}$ denotes the offset of activation function of the atom neuron $A_{q_i^j}$. Since we have chosen the Avg-Sigmoid function family, the outputs of aggregation neurons are further calculated as
$$\textit{Agg}_{(h \leftarrow b_1 \wedge \dots \wedge b_{n_k},w_k)}^{q_i^j} = \frac{1}{l}\sum_{m=1}^{l} R_{q_i^j \leftarrow b_1\theta_m \wedge \dots \wedge b_{n_k}\theta_m}$$
where $R_{q_i^j \leftarrow b_1\theta_m \wedge \dots \wedge b_{n_k}\theta_m}$ denotes outputs of respective input rule neurons formed from all different groundings (substitutions $\theta_m$) of the rule $R_{h\theta_m \leftarrow b_1\theta_m \wedge \dots \wedge b_{n_k}\theta_m}$ where $h\theta_m = q_i^j$. The output of the rule neurons can finally be calculated as
$$R_{q_i^j \leftarrow b_1\theta_m \wedge \dots \wedge b_{n_k}\theta_m} = \textit{sigm} \left( \left(\sum_{o=1}^{n_k} A_{b_{o}^j\theta_m}\right) - n_k \right)$$
where 
$A_{b_{o}^j\theta_m}$ denotes output of another (regular) atom neuron from the lower layers of the ground network $\overline{\mathcal{N}\cup \mathcal{E}^j}$ corresponding to one of the ground body literals $b_{o}\theta_m$ of the respective ground rule $q_i^j \leftarrow b_1\theta_m \wedge \dots \wedge b_{n_k}\theta_m$. The~calculation of $A_{b_{o}^j\theta_m}$ can further be carried out in a recursive manner until the fact neurons $F_{(h,w)}$ are reached with fixed constant values defined by $\mathcal{E}$ (or possibly $\mathcal{N}$). We note that the whole evaluation composed of differentiable functions and the gradient $\frac{\partial J(\mathcal{Q})}{\partial w_h}$ can thus be calculated using regular chain rule.

\end{example}

Moreover, the weights from $\mathcal{N}$ can be repeated multiple times within a single $\overline{\mathcal{N}\cup \mathcal{E}^j}$, but since recursion is not allowed, the same weight can appear at most once on any simple path from a fact neuron to an atom neuron. Therefore it is possible to learn the weights using conventional online stochastic gradient descent algorithm\footnote{Learning is slightly more complicated for LRNNs with the Max-Sigmoid family of activation functions because the $\max$ operator introduces non-differentiable points to the optimization problem.}, except that the increments for the shared weights must be accumulated, which is a simple consequence of linearity of partial differentiation. The same principle is exploited e.g.\ in learning of convolutional neural networks (Example~\ref{ex:CNN}).

\begin{remark}
Let us consider a ground $\overline{\mathcal{N}\cup \mathcal{E}^j}$ as a regular feed forward neural network $N_j$ with some weights $w_k \in \mathcal{W}^j$ in the network being shared, i.e. bound to the same value, with the restriction that each particular weight $w_k$ appears at most once on any simple path from input $e_j$ to output $y_j$. Let the activation functions of layers $l$ of $N_j$ be $f^l \in \mathcal{F}^j$ from some set of differentiable functions. Let further $w_k^i$ denote particular occurrences of some shared weight $w_k$, then we might express the output of the network as
$$ y_j =  f^1\left( \ldots + w_k^a f^2\left(\ldots\right) + \ldots + w_m f^2\left(\dots + w_k^b f^3\left(\ldots\right)+\ldots\right) + \ldots\right) $$
where "$\ldots$" correspond to expressions with no $w_k$ occurrence. Considering each $w_k$ occurrence separately as an independent variable, we have
$$ \frac{\partial y_j}{\partial w_k^a} = \frac{\partial \left(f^1\left( w_k^a f^2(\ldots) \right)\right)}{\partial w_k^a} = f^1{'}(\ldots) f^2(\ldots) $$
$$ \frac{\partial y_j}{\partial w_k^b} = \frac{\partial\left(f^1 \left(w_m f^2 \left(w_k^b f^3 \left(\ldots\right)\right)\right)\right)}{\partial w_k^b} = f^1{'}(\ldots) \: w_m f^{2}{'}\left(\ldots\right) f^{3}\left(\ldots\right) $$
Considering all occurrences of $w_k^i$ as a single variable $w_k$, we have
$$
\frac{\partial y_j}{\partial w_k} = \frac{\partial \left(f^1\left( w_k f^2(\ldots) + w_m f^2 \left(w_k f^3 \left(\ldots\right)\right)\right)\right)}{\partial w_k} = f^{1}{'}\left(\ldots\right) \left(f^2\left(\ldots\right) + w_m f^{2}{'}\left(\ldots\right) f^{3}\left(\ldots\right) \right)
$$
i.e., we see that $\frac{\partial y_j}{\partial w_k} = \frac{\partial y_j}{\partial w_k^a} + \frac{\partial y_j}{\partial w_k^b}$ which follows also directly from additivity of the differentiation operator (keeping in mind that there is only one occurrence of $w_k$ on any simple path from an atom neuron to a fact neuron). Therefore gradient can be computed for the ground neural networks created from a given LRNN in the standard way and then the components corresponding to a particular weight $w_k$ can be accumulated.
\end{remark}

Specifically, our weight-learning algorithm works as follows. First, it grounds the given LRNN $\mathcal{N}$ w.r.t.\ every example $\mathcal{E}^j$ from the dataset which gives it a set of ground neural networks $\overline{\mathcal{N}\cup \mathcal{E}^j}$ with shared weights (it keeps the information about the origin of each weight so that it could update the respective weights in the template in each step of the iteration). It then iterates over the ground networks in a random order, computes gradient of the error function for the current particular example given the current weights in the template, updates the weights accordingly and continues iterating these steps (i.e., the standard stochastic gradient descent procedure). In order to reduce the risk of getting stuck in poor quality local optima, we also employ a restart strategy for this algorithm.
 
\section{Related Work}\label{sec:related_work}

The main inspiration for the work presented in this paper are lifted graphical models such as Markov logic networks \citep{MLN} or Bayesian logic programs \citep{BLP}. However, none of these existing lifted graphical models is particularly well suited for learning parameters of latent relational structures. Our approach is also generally related to prior art in combining logical rules with neural networks, also known as neural-symbolic integration~\citep{Garcez2012}, such as in the KBANN system. While the KBANN \citep{KBANN} also constructs the network structure from given rules, these rules are propositional rather than relational and do not serve as a lifted template. Therefore it is impossible to learn relational latent structures such as soft clustering of first-order-logic constants. A more recent system CILP++\citep{CILP} utilizes a relational representation, which is however converted into a propositional form through a propositionalization technique~\citep{propos}. This again means that latent relational structures such as those exemplified in Section \ref{sec:concepts} cannot be learned by CILP++ either. 
A somewhat more closely related paper on FONN~\citep{Botta} also designs a technique forming a network from relational rule set, however this rule set is flat, producing only 1-layer (shallow) networks in which relational patterns are not hierarchically aggregated.
While there are many other approaches of neural-symbolic integration aiming at relational (and first-order) representations~\citep{Bader2005a}, e.g.\ based on the CORE method~\citep{Holldobler1999}, they typically search for a uniform model of the logic program in scope and thus principally differ from the presented lifted modeling approach.

While standard feed-forward neural networks can be seen as a special case of LRNNs, since any such a fixed neural architecture can be encoded in a corresponding ground rule set with respective activation functions, a salient aspect of our method is that it allows for learning from structured (relational) examples, rather than just attribute vectors. There has been previous work on adapting neural networks to cope with certain facets of relational representations. For example, extension to multi-instance learning was presented in~\citep{Ramon}. A similarly directed work~\citep{Blockeel} facilitated aggregative reasoning to process sets of related tuples from relational database as a sequence through recurrent neural network structure, which was also presented for more general structures in~\citep{Scarselli2009a}. These approaches are principally different from the presented method as they do not follow the lifted modeling strategy to cope with variations in structure of relational samples.
More loosely related works arise also in the neural networks community, where various recursive auto-encoders based on the idea of ``reduced descriptions''~\citep{Hinton1990} are trained to encode structured data. Another line of work are convolutional neural networks~\citep{LeCun1998} and techniques of indirect encoding~\citep{Clune}, exploiting patterns and regularities in neural connections to create more compressed representations of large neural networks. However, these approaches are still geared towards learning from fixed-length propositional rather than relational data.

\section{Experiments}\label{sec:experiments}

In this section we describe experiments performed on 78 datasets of organic molecules: Mutagenesis dataset~\citep{mutagenesis}, four datasets from the predictive toxicollogy challenge and 73 NCI-GI datasets \citep{ncigi}. The Mutagenesis dataset contains 188 molecules with labels denoting their mutagenicity. A number of the results published on the mutagenesis dataset use extended set of features, providing additional expert knowledge on relational properties of molecules, degrading the role of learning capabilities in relational models. We do not use any of the extra features as we utilize only atom-bond information. The predictive toxicology challenge dataset (PTC)~\citep{PTC} is composed of four datasets of molecules labeled by their toxicity for female rats (fr), mouse (fm) and male rat (mr) and mouse (mm). Each of the NCI-GI datasets contains several thousands of molecules labeled by their ability to inhibit growth of different types of tumors. We compare performance of LRNNs to state-of-the-art relational learners kFOIL \citep{kFOIL} and nFOIL \citep{nFOIL}, where kFOIL combines relational rule learninng with support vector machines and nFOIL combines relational rule learning with naive Bayes learning.

For LRNNs we use a simple hand-crafted template which is based on the idea of implicit soft clustering described in Section \ref{sec:soft_clustering} and is principally identical to the template discussed in Figure \ref{fig:template}. The template defines 3 predicates for clusters of atom types and 3 predicates for clusters of bond types. The three predicates representing atom-type clusters are composed of exhaustive lists of atom types occurring in the datasets, e.g.\ $w_1^{(1)} : \textit{atgr1}(X) \leftarrow \textit{o}(X)$, $w_1^{(1)} : \textit{atgr1}(X) \leftarrow \textit{br}(X)$, $\dots$ and similarly the predicates representing bond-type clusters are composed of exhaustive lists of bond types occurring in the datasets. These predicates are then used in definitions of predicates for different types of small chains of atoms of length 3, e.g.\ $\textit{chain1}$ $\leftarrow$ $\textit{atgr1}(X)$ $\wedge$ $\textit{bond}(X,Y,B1)$ $\wedge$ $\textit{atgr1}(Y)$ $\wedge$ $\textit{bond}(Y,Z,B2)$ $\wedge$ $\textit{atgr2}(Z)$ $\wedge$ $\textit{bondgr1}(B1)$ $\textit{bondgr2}(B2)$. These are finally used to define the target predicate, e.g.\ {\em toxic}. Using such a generic template for all the datasets, we make sure that there is no additional expert knowledge involved~\footnote{I.e., the template does not relate to any specific property of molecules and might be as well used for other classification tasks, too.}. The idea is that in the process of learning, useful latent relational concepts are created within the neural network by the means of weight adaptation rather than by explicit enumeration, in contrast to propositional approaches and ILP~\citep{Raedt}. Indeed, none of the rules used in this template is useful on itself for prediction as a hard logic rule without weight adaptation.

\begin{figure}[t]
\centering
\includegraphics[width=\textwidth, height=8cm]{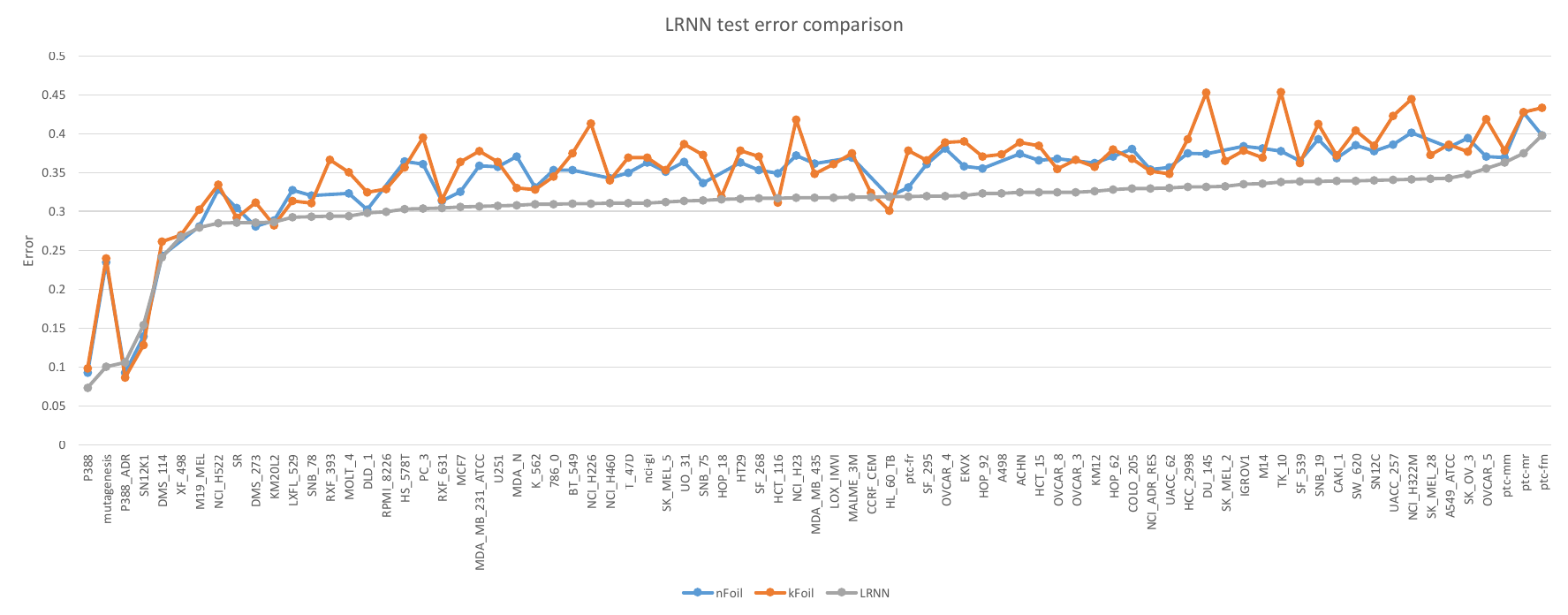}
\caption{Prediction errors of LRNNs, kFOIL and nFOIL measured by cross-validation on 78 datasets of organic molecules.}
\label{fig:results}
\end{figure}

To set the parameters of LRNNs we use the empirical risk minimization principle on the training cross-validation folds to select the parameters such as step size, restarts, number of iterations, etc. This way we obtain unbiased estimates of performance of our methods since test data is never involved in parameter selection. The time for training a LRNN was in the order of few hours for the larger NCI-GI datasets. The results of the experiments are summarized in Figure \ref{fig:results}. LRNNs perform clearly the best of the algorithms in terms of accuracy as they have lower prediction error than kFOIL and nFOIL on significant majority of datasets. We also tried to compare LRNNs with another recent algorithm combining logic and neural networks, called CILP++ \citep{CILP}, but we didn't find it to perform well on our relational datasets as we were not able to obtain, using CILP++, accuracy significantly higher than simple majority class error on any of the datasets\footnote{While relatively reasonable results for Mutagenesis were reported in \citep{CILP}, the expert-knowledge attributes were used in the experiments reported therein, which might explain the discrepancy between the results.}.

\begin{figure}[h!]
\centering
\resizebox{\textwidth}{3cm}{
\begin{minipage}[b]{0.5\linewidth}
\includegraphics[width=\textwidth]{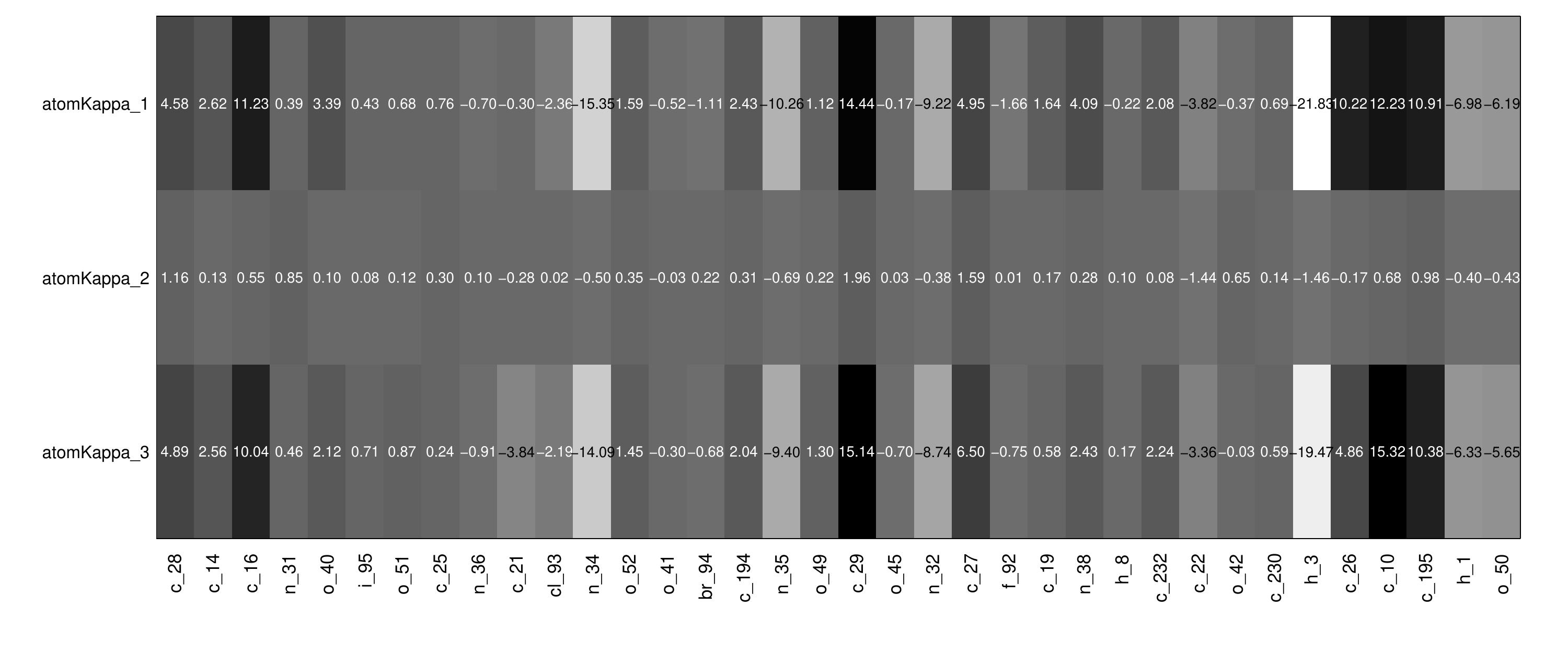}
\end{minipage}
\hspace{0cm}
\begin{minipage}[b]{0.5\linewidth}
\includegraphics[width=\textwidth]{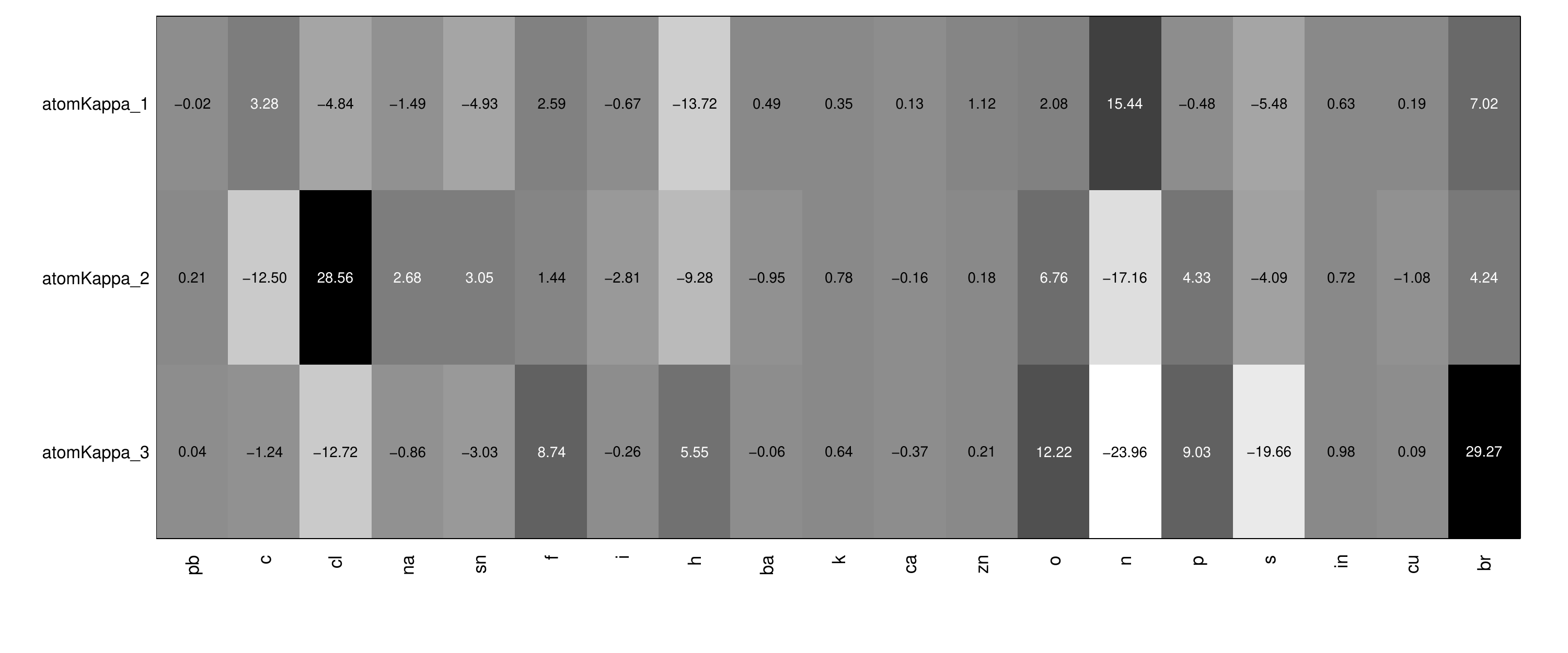}
\end{minipage}}
\caption{Visualization of latent concepts demonstrated through LRNN's weights of rules defining particular groups of atoms ($Kappa_{1..3}$) when learned in the Mutagenesis dataset (left) and in the PTC-mr datasets. Lighter colors denote lower and darker colors higher weights, respectively.}
\label{fig:conceptsMuta}
\end{figure}

For demonstration, we provide visualization of the latent grouping (clustering) LRNN layers for the Mutagenesis and for the PTC-mr datasets in Fig.~\ref{fig:conceptsMuta}. It is apparent from the learned weights in figures that the hidden layers are indeed learning useful latent groupings of atom types. It is interesting to note that on the Mutagenesis dataset, one of the learned groupings of atom types gives all atoms almost the same weight, which actually makes sense because it corresponds to a ``wild-card'' atom type. On the other hand, no similar behavior is typically found for the other datasets, which we have checked for different seeds of the PRNG used for initialization of weights.

In order to test the modeling concept described in Section \ref{sec:other_concepts}, we performed an additional experiment with the Mutagenesis dataset. We used almost exactly the same template as in Example \ref{ex:other_concept} but instead of ring structures we used chains of varying lengths (up to 5 atoms). We trained the resulting LRNN to optimize the template's weights, however here we were more interested in extracting the learned patterns. We determined the chains of atoms which gave the highest output for the learned latent predicates. We obtained the following atom chain structures: 
C-C-F,
N-O,
C-Cl,
C-Br,
C-C-O,
O-N-C. At least some of these structures appear to be directly relevant for the mutagenicity as they contain organic structures containing halogen atoms (Br, F and Cl). The other structures may be relevant to mutagenicity in combination with other structures.

\section{Conclusions}

In this paper, we have introduced a method combining relational-logic representations with feedforward neural  networks. The introduced method is close in spirit to lifted graphical models as it can be viewed as providing a lifted model for construction of ground neural networks. The performed experiments indicate that it is possible to achieve state-of-the-art predictive accuracies by weight learning with very generic templates and that it is able to induce notable auxiliary concepts. There are many directions for future work, including structure learning, transfer learning or studying different collections of activation functions. An important future direction is also the question of extending LRNNs to support recursion.

\subsubsection*{Acknowledgments} GS and F\v{Z} are supported by Cisco sponsored research project ``Modelling network traffic with relational features''. While with CTU, OK was supported by the Czech Science Foundation through project P202/12/2032 and now by a grant from the Leverhulme Trust (RPG-2014-164).

\bibliography{main}

\end{document}